\documentclass[review,12pt]{elsarticle}

\usepackage{amssymb}
\usepackage{amsmath}
\usepackage{url}
\usepackage{subcaption}
\usepackage{booktabs}
\usepackage{multirow} 
\usepackage[T1]{fontenc}
\usepackage{hyperref}

\journal{}

\begin{document}

\begin{frontmatter}



\title{Enhancing Cryptocurrency Sentiment Analysis with Multimodal Features}


\author[inst1]{Chenghao Liu\corref{cor1}}
\ead{chenghao.liu@auckland.ac.nz}

\author[inst1]{Aniket Mahanti}
\ead{a.mahanti@auckland.ac.nz}

\author[inst2]{Ranesh Naha}
\ead{ranesh.naha@qut.edu.au}

\author[inst3]{Guanghao Wang}
\ead{guanghao.wang@auckland.ac.nz}

\author[inst3]{Erwann Sbai}
\ead{e.sbai@auckland.ac.nz}

\affiliation[inst1]{organization={Department of Computer Science, The University of Auckland},
            city={Auckland},
            country={New Zealand}}

\affiliation[inst2]{organization={School of Information Systems, Queensland University of Technology},
            city={Brisbane},
            state={Queensland},
            country={Australia}}

\affiliation[inst3]{organization={Department of Economics, The University of Auckland},
            city={Auckland},
            country={New Zealand}}

\begin{abstract}
As cryptocurrencies gain popularity, the digital asset marketplace becomes increasingly significant. Understanding social media signals offers valuable insights into investor sentiment and market dynamics. Prior research has predominantly focused on text-based platforms, such as Twitter\footnote{Renamed as X in July 2023; however, we continue to refer to it as Twitter due to the widespread recognition and entrenched nature of the original brand}. However, video content remain underexplored, despite potentially containing richer emotional and contextual sentiment that is not fully captured by text alone. In this study, we present the multimodal analysis comparing TikTok and Twitter sentiment, using large language models to extract insights from both video and text data. We investigate the dynamic dependencies and spillover effects between social media sentiment and cryptocurrency market indicators. Our results reveal that TikTok’s video-based sentiment significantly influences speculative assets and short-term market trends, while Twitter’s text-based sentiment aligns more closely with long-term dynamics. Notably, the integration of cross-platform sentiment signals improves forecasting accuracy by up to 20\%.
\end{abstract}

\begin{graphicalabstract}
\end{graphicalabstract}

\begin{highlights}
\item We introduce a framework for analyzing TikTok video-based sentiment compared to Twitter text-based sentiment in cryptocurrency sentiment analysis.
\item  We find that TikTok sentiment particularly influences short-term speculative trends, while Twitter sentiment aligns better
with long-term market dynamics. 
\end{highlights}


\end{frontmatter}



\section{Introduction}
\label{sec1}

Since Bitcoin's emergence in 2008, it has pioneered a new class of digital assets, reshaping the financial landscape. By operating without traditional intermediaries, Bitcoin relies on cryptographic technology to maintain a decentralized structure free from direct governmental or central bank control. This positions cryptocurrencies as potential portfolio diversifiers, especially during economic instability, given their reduced linkage to conventional economic factors \cite{caferra2021raised}. The cryptocurrency market has rapidly grown, reaching a market capitalization of \$3.65 trillion by December 2024\footnote{\url{https://coinmarketcap.com/charts/}}, driven by both Bitcoin and specialized altcoins. Recently, the U.S. government's creation of a strategic cryptocurrency reserve including Bitcoin, Ethereum, XRP, Solana, and Cardano has further emphasized their economic significance, boosting market optimism and price appreciation\footnote{\url{https://www.bbc.com/news/articles/cn0jgggd7r4o}}. This growth coincides with increased academic interest, with cryptocurrency publications in top-tier journals rising approximately 45.92\% annually \cite{pevciulis2024forecasting}. Additionally, global cryptocurrency adoption surpassed 617 million users by 2024\footnote{\url{https://www.statista.com/statistics/1202503/global-cryptocurrency-user-base/}}, highlighting their widespread acceptance and integration into financial practices worldwide.

In parallel with these technological and economic developments, the interplay between cryptocurrency and political narratives has become increasingly pronounced. For instance, prior to President Donald Trump second inauguration, he leveraged social media platforms to promote a meme coin known as \$TRUMP, encapsulating the fusion of political messaging, digital branding, and speculative investment strategies. The coin, which rapidly ascended to become one of the top 20 cryptocurrencies by market capitalization\footnote{\url{https://coinmarketcap.com/currencies/official-trump/}}, serves as a striking example of how cultural relevance and social momentum can drive market dynamics, often overshadowing traditional financial or technological fundamentals \cite{yousaf2023connectedness,nani2022doge,huang2024memes}. Such politically driven promotions expose retail investors to heightened risks, including market volatility and speculative bubbles, potentially resulting in significant financial losses. For regulators and policymakers, understanding these trends is crucial for establishing informed regulatory frameworks that ensure market transparency, investor protection, and overall financial stability. Investors and market participants gain from recognizing the potential vulnerabilities, allowing them to implement more robust risk management strategies.

Beyond meme coins and political endorsements, cryptocurrencies differ significantly from traditional financial assets due to their lack of intrinsic valuation metrics, such as cash flows or dividends, complicating fair-value assessments and enhancing their speculative nature \cite{bhambhwani2019fundamentals,biais2023equilibrium,delfabbro2021psychology,sockin2023model,yermack2024bitcoin,baur2018bitcoin}. Given this uncertainty, investor sentiment is crucial in shaping cryptocurrency market behavior. Literature on sentiment's role typically emphasizes three areas: (1) emotions, perceptions, and social trends as primary drivers of returns \cite{akyildirim2021investor,liu2021risks,smales2022investor}; (2) social media's influence through investor disagreements, crowd wisdom, influential figures, and echo chambers as key volatility sources \cite{cookson2023echo,cookson2020don,grennan2021fintechs}; and (3) empirical evidence from surveys, media coverage, and social analytics demonstrating sentiment's predictive power for short-term returns and trading opportunities \cite{stambaugh2012short,chen2014wisdom,zhou2018measuring}. Research on cryptocurrencies specifically indicates influencer-generated content on platforms like YouTube significantly affects prices and volumes, particularly for low-cap assets, while aggregate social media activity broadly shapes market trends \cite{moser2023should,meyer2024testing,naeem2021predictive,lath2022impact}. Yet, gaps remain regarding how specific content features, such as sentiment, tone, and thematic focus—influence cryptocurrency market dynamics.

Our paper is guided by the following research questions (\textbf{RQs}):
\begin{enumerate}
    \item What is the relationship between video-based sentiment from TikTok and cryptocurrency price returns and volume changes, compared to text-based sentiment from Twitter?
    \item How can integrating multimodal sentiment from TikTok with text-based sentiment from Twitter enhance the prediction of cryptocurrency market trends?
\end{enumerate}

This study makes three contributions. First, we introduce a framework for analyzing TikTok video-based sentiment compared to Twitter text-based sentiment. Second, we find that TikTok sentiment drives short-term speculation, whereas Twitter reflects long-term trends. Third, integrating these sentiments improves forecasting accuracy and clarifies how social media influences investor behavior. Our findings show combining TikTok and Twitter sentiment enhances crypto return and volume forecasts by up to 20\%, with TikTok alone improving short-term Dogecoin predictions by 35\%. Stablecoins respond less to sentiment due to their pegged values. TikTok and Twitter receive the most volatility shocks (1. 90\%, 4. 65\%), while Dogecoin strongly transmits volume spillovers (83.90\%).

Our results demonstrate that integrating TikTok and Twitter sentiment signals significantly enhances forecasting accuracy for cryptocurrency price returns and trading volumes. We find that TikTok sentiment particularly influences short-term speculative trends, while Twitter sentiment aligns better with long-term market dynamics. Additionally, Dogecoin is identified as the primary transmitter of volume spillovers, whereas stablecoins exhibit limited responsiveness to social media sentiment.

\section{Related Work}
\label{sec2}
This section introduces literature on sentiment analysis methods and spillover effects in cryptocurrency markets, emphasizing the predictive role of social media sentiment and highlighting the underexplored potential of video-based platforms.

\subsection{Methodologies for Sentiment Analysis and Forecasting}
Advancements in natural language processing have significantly propelled Financial Sentiment Analysis (FSA), covering tasks at document, paragraph, sentence, and aspect levels \cite{mao2021bridging,zhu2024neurosymbolic,du2024financial}. Domain-specific transformers like FinBERT have enhanced FSA, yet the use of autoregressive decoders such as GPT remains underexplored, with recent studies beginning to address this gap through approaches like zero-shot prompting \cite{fatemi2023comparative,zhang2023enhancing} and by integrating retrieval-augmented mechanisms \cite{zhang2023instruct,hu2023whetting}. In parallel, multimodal methods that combine vocal, textual, and visual cues have shown promise in financial forecasting, although the vision modality is still largely underutilized despite its potential \cite{sawhney2020multimodal,mathur2022monopoly,cao2022ai}. Despite these advances, there is a notable gap in research on how text-based and multimodal sentiment interact, particularly regarding their influence on cryptocurrency price returns and trading volumes across diverse social platforms like TikTok and Twitter.

\subsection{Information Spillover in Cryptocurrency Market}
Numerous studies document information spillover in cryptocurrency markets. Fry and Cheah \cite{fry2016negative} observed spillover from Ripple to Bitcoin, while Corbet et al. \cite{corbet2018datestamping} and Kyriazis \cite{kyriazis2019survey} highlighted interlinkages among Bitcoin, Ethereum, Litecoin, and Ripple, with Yi et al. \cite{yi2018volatility} noting that high-capitalization cryptocurrencies tend to propagate volatility shocks. Further, Ji et al. \cite{ji2019dynamic} and Aslanidis et al. \cite{aslanidis2021cryptocurrencies} confirmed that shocks from dominant cryptocurrencies, especially Bitcoin, serve as net transmitters of market disturbances. Social sentiment also plays a critical role, as Aste \cite{aste2019cryptocurrency} and Bourghelle et al. \cite{bourghelle2022collective} demonstrated a causal relationship between sentiment and price fluctuations. Yet, significant gaps remain in understanding how these dynamics operate across platforms like TikTok and Twitter, and in assessing the impact of non-textual, video-based sentiment on market interconnectedness.

\subsection{Comparison with Prior Work}
Existing research on cryptocurrency sentiment has focused mainly on text based data from platforms such as Twitter or Reddit or on aggregated effects across social platforms, often neglecting the unique features of short form video content. Traditional approaches leverage lexicon or machine learning based sentiment analysis \cite{georgoula2015using}, whereas our work bridges this gap by incorporating both text based (Twitter) and video based (TikTok) sentiment signals. Unlike earlier studies that concentrate on limited time frames or single modalities, we employ multimodal large language models (mLLMs) to combine text, audio, and video features into a holistic framework. This approach captures subtleties such as tone of voice and visual cues, offering richer insights than text only analysis. mLLMs are capable of processing and combining video information such as tone of voice, facial expressions, and background visuals with traditional text data \cite{miah2024multimodal}. Recent studies on large language models show promise in financial sentiment analysis but rarely address the complexities of non textual data or evaluate how multiple platforms jointly affect market behavior \cite{fatouros2023transforming,zhang2023instruct}. By merging multimodal TikTok sentiment with Twitter data, our method reveals platform specific dynamics, demonstrating how visually driven short form content can trigger short lived price momentum while Twitter based discussions may produce longer term signals, and thus extends sentiment analysis beyond text to better predict cryptocurrency price movements and trading volumes.

\begin{figure*}[ht]
    \centering
    \scalebox{1.05}{ 
        \includegraphics[width=\linewidth]{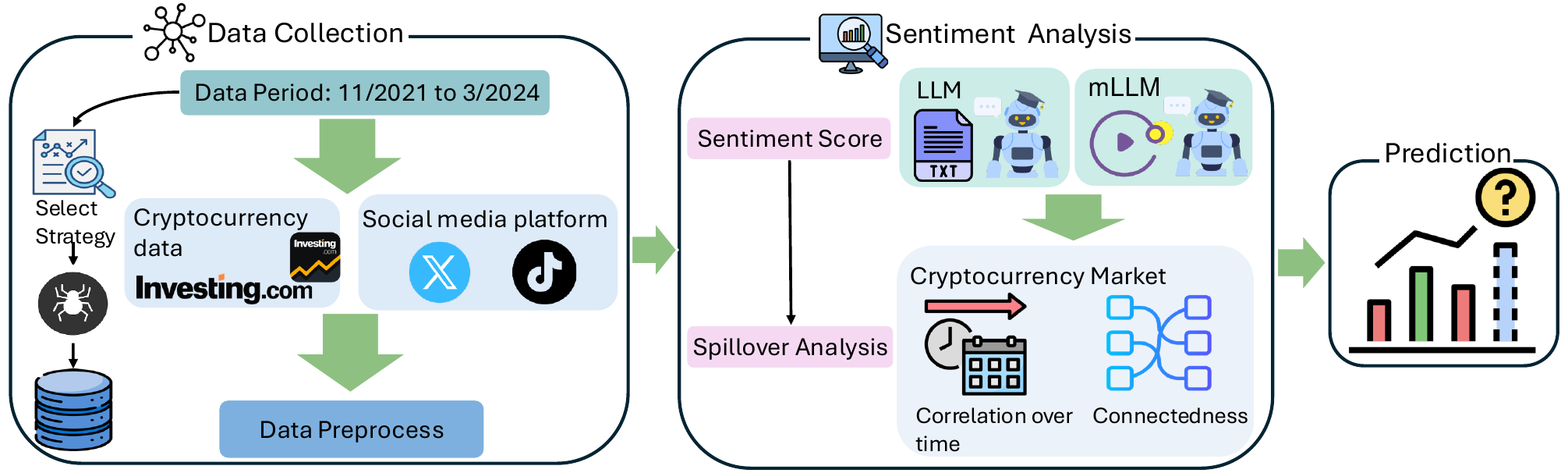}
    }
\caption{Workflow for Sentiment Analysis and Forecasting.}    
\label{fig:workflow}
\end{figure*}

\section{Methodology}
\label{sec3}
\subsection{Data Collection}
The workflow pipeline Figure \ref{fig:workflow} shows the data collection process, which consists of three key components: Twitter data, TikTok data, and cryptocurrency price and volume data. These datasets are collected through Tweepy\footnote{\url{https://www.tweepy.org/}}, the TikTok API\footnote{\url{https://github.com/Evil0ctal/Douyin_TikTok_Download_API}}, and Investing.com\footnote{\url{https://www.investing.com/crypto/currencies}}, respectively. Each component captures a distinct aspect of cryptocurrency sentiment and market trends, offering a multi-dimensional perspective for analysis.
Twitter captures public sentiment, TikTok focuses on influencer video content, and price data provides market activity. Following Hardle et al. \cite{hardle2020understanding}, cryptocurrencies are classified into Gold 2.0 (BTC), Altcoins (ETH, SOL, DOG, XRP) and Stablecoins (USDC, USDT). Basic preprocessing, such as removing unnecessary elements and standardizing text, was applied to refine the Twitter datasets for sentiment analysis.

\textbf{Twitter Data:} From November 8, 2021 to March 13, 2024, 519,208 public Tweets related to cryptocurrency were collected as JSON line objects using a Tweepy-based web crawler. Tweets containing keywords such as “BTC”, “\$BTC”, or “Bitcoin” were automatically stored, providing a structured dataset for further analysis, with Twitter widely recognized in financial market studies \cite{saleem2024twitter,ante2023elon,bouteska2023effect}.

\textbf{TikTok Data:} A dataset of 15,348 TikTok video posts was collected using a multi step method adapted from Cheng and Li \cite{cheng2024like}. TikTok was chosen for its popularity and influence among young audiences \cite{cervi2021tik,jagolinzer2024market}. The top 10 most-followed cryptocurrency influencers were identified (see Table \ref{tab:tik_influencers}) and their video posts retrieved via the TikTok API. From these videos, 50 unique hashtags were extracted from closed captions and the 37 most viewed were selected; only English-language videos with subtitles were retained, with transcription enhanced using Whisper \cite{radford2023robust}. \\

We identified the top 10 most-followed cryptocurrency influencers on TikTok to ensure a comprehensive representation of high-engagement content. Table \ref{tab:tt-account} shows these influencers' details. We then utilized a web scraping tool based on the TikTok API to retrieve all video posts shared by these influencers, focusing on content related to cryptocurrency topics, trends, and discussions.

\begin{table}[htbp]
\centering
\caption{Top 10 Most-Followed Cryptocurrency Influencers on TikTok}
\label{tab:tik_influencers}
\resizebox{\columnwidth}{!}{%
\begin{tabular}{lccccc}
\hline
\textbf{Account Name} & crypto.lucho & CryptoMasun & CryptoProjects & thecryptonetwork & cryptosanti \\
\hline
\textbf{Number of Followers} & 2.7M & 1.4M & 1.3M & 1.0M & 914K \\
\hline
\textbf{Account Name} & cryptorobles & suppomancrypto & Virtual Bacon & thecryptohippie & Cryptowendyo \\
\hline
\textbf{Number of Followers} & 886K & 344K & 338K & 288K & 277K \\
\hline
\label{tab:tt-account}
\end{tabular}%
}
\end{table}

\textbf{Cryptocurrency Data:} Daily closing prices (in USD) and trading volumes for the top 7 cryptocurrencies (BTC, ETH, SOL, DOG, XRP, USDC, USDT) were sourced from Investing.com. To ensure quality, only coins with at least 1,000 observations were included, representing over 90\% of the market's total capitalization. Notably, Bitcoin reached a new all time high in March 2024, and one representative coin from each group was used for dependency analysis and forecast modeling.

For each group, one representative coin was chosen to conduct dependency analysis between sentiment signals and forecast modeling, providing insights into the differing dynamics across cryptocurrency categories. A data summary is presented in Table \ref{tab:datacol}.

\begin{table}[h!]
\centering
\caption{Data Summary for TikTok and Twitter}
\label{tab:datacol}
\resizebox{\columnwidth}{!}{%
\begin{tabular}{lccccc}
\toprule
\textbf{Platform} & \textbf{Period} & \textbf{Number of data} & \textbf{Average likes} & \textbf{Average comments} & \textbf{Average shares} \\
\midrule
TikTok  & 2021.11--2024.03 & 15,348  & 1,350 & 210 & 90 \\
Twitter & 2021.11--2024.03 & 519,208 & 240   & 35  & 15 \\
\bottomrule
\end{tabular}%
}
\label{tab:datacol}
\end{table}

\subsection{Sentiment Analysis Framework}
\subsubsection{Twitter Sentiment Analysis}
\begin{figure}[ht]
    \centering
    \scalebox{0.75}{
        \includegraphics[width=\linewidth]{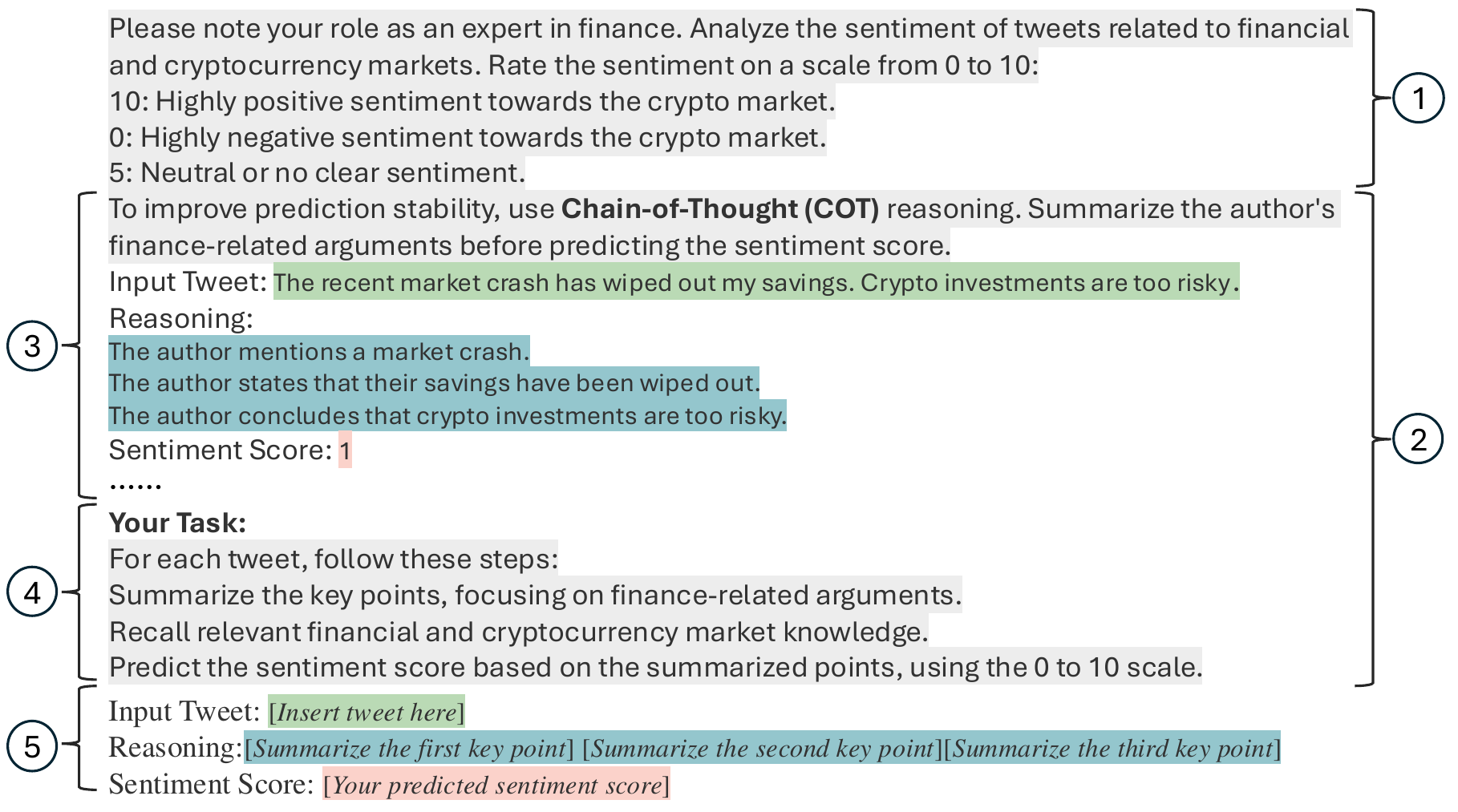}
    }
    \caption{Llama3 sentiment analysis prompt template}
    \label{fig:prompt-template}
\end{figure}

We annotate tweet sentiment using the Meta-Llama-3-8B model \cite{touvron2023llama}. Although it is the smallest in the Llama series, its performance exceeds random baselines and open source models while matching closed source alternatives. Our objective is to extract the author's financial outlook on cryptocurrencies by adapting the task to capture expectations about market trends, as shown in Figure \ref{fig:prompt-template}. This task simulates the target domain, presents multiple input-output examples, and predicts a sentiment score. Our preliminary study indicates that the prompt yields reasonable results but is unstable and sensitive to wording and demonstration order \cite{min2022rethinking}.

To mitigate this instability, we incorporate chain of thought reasoning \cite{wei2022chain} to guide the model in summarizing finance related arguments and recalling domain knowledge. We use temperature sampling \cite{ackley1985learning} to generate multiple reasoning paths when users present conflicting arguments. Following Wang et al. (2024) \cite{wang2024interrelations}, we adopt a sentiment scoring range from 0 to 10, where 0 indicates strong confidence in a market decline, 10 indicates strong confidence in a market rise, and 5 is neutral. The average daily sentiment inclination score, denoted as $TSI_t$, is computed as
\[
TSI_t = \frac{1}{n}\sum_{k=1}^{n} S_k,
\]
with $n$ representing the number of tweets on day $t$ and $S_k$ the sentiment score for the $k$th tweet.

\subsubsection{Tiktok Data Sentiment Analysis}
\begin{figure}[htbp]
    \centering
    \scalebox{0.65}{
    \includegraphics[width=\linewidth]{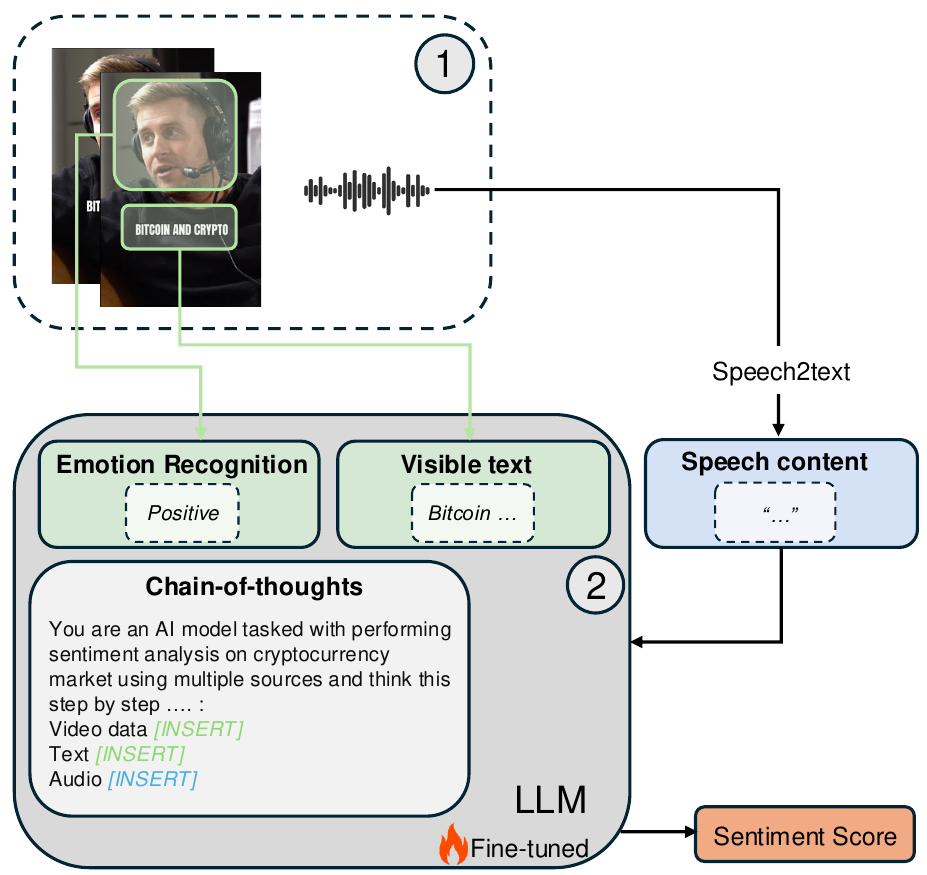}
    }
    \caption{Cross-Modal Sentiment Analysis Pipeline for TikTok Video Data}
    \label{fig:video-archicture-sec4}
\end{figure}

Our process for TikTok sentiment analysis comprises two main components: the Cross Modal Alignment Module and the Answer Module. In the Cross Modal Alignment Module, a long video is segmented into non overlapping short clips by dividing the sequence of frames into chunks that fit within the MiniGPT4-Video context limit. For each clip, MiniGPT4-Video generates a concise summary that captures key visual and contextual information. These summaries, along with any associated subtitles, are then encoded into fixed dimensional embeddings using OpenAI's \texttt{text-embedding-3-small} model. The embeddings represent the semantic content of each clip, and together they form a comprehensive representation of the video content. The Answer Module then takes these clip embeddings along with an encoded user query to generate a final response via Llama2-chat \cite{touvron2023llama}, effectively linking the visual and textual cues to sentiment outcomes. Figure \ref{fig:video-archicture-sec4} illustrates this cross modal sentiment analysis pipeline.

To further enhance sentiment detection, we fine tune the entire model using LoRA \cite{hu2021lora} with a high quality video sentiment analysis dataset. This fine tuning process refines the model's ability to interpret subtle cues in video content and generate accurate sentiment responses. In addition, we incorporate the MEAD \cite{wang2020mead} benchmark dataset, which is widely used for multimodal sentiment analysis, to improve performance and ensure robustness across varied sentiment expressions. Together, these steps allow our approach to effectively capture and analyze sentiment in TikTok videos by integrating visual, audio, and textual information.

\subsection{Spillover Analysis}
After data collection and preprocessing, we analyze how social media sentiment interacts with cryptocurrency prices and volumes. We employ Word LDA for topic identification, Rolling Window Correlation for time varying relationships, and a DCC-GARCH based connectedness approach to evaluate shock transmission among variables.

\subsubsection{Topic Visualization}
We use the LDA model \cite{blei2007correlated} to cluster words and uncover patterns in TikTok and Twitter data. Text is transformed into TF-IDF features and grouped with K-means clustering to determine the optimal number of topics based on word co-occurrence \cite{jeyaraj2020evolution}. This reveals differences in topic prevalence and sentiment expression across platforms.

\subsubsection{Rolling Window Correlation}
A sliding window technique is applied to compute time varying correlations between cryptocurrency prices, volumes, and sentiment data \cite{bouri2017hedge, corbet2018exploring}. By using window sizes of 7, 14, 32, 64, 128, and 256 days with a one day step, we capture both short term and long term trends, balancing local details with overall patterns.

\subsubsection{DCC-GARCH Based Connectedness Approach}
This method models time varying conditional variances and covariances, eliminating the need for fixed window sizes \cite{engle2002dynamic}. We begin with a generalized forecast error variance decomposition (GFEVD) to measure how much each variable influences others \cite{gabauer2020volatility}. Directional connectedness measures (TO and FROM) are computed to assess shock transmission, and their difference indicates whether a variable is a net transmitter or receiver. Finally, the Total Connectedness Index (TCI) represents the overall contribution of each variable’s forecast error variance.

\subsection{Prediction Task}
Integrating sentiment data from Twitter and TikTok into forecasting models captures market sentiment dynamics and their impact on cryptocurrency movements \cite{deng2023llms, smailovic2013predictive, xing2020high}. Our objective is to evaluate the effectiveness of this sentiment data in predicting cryptocurrency price returns and volume changes over various forecast horizons. Price return measures the relative change in price over a period, while volume change reflects fluctuations in trading activity. For each task, the model is trained to forecast the cryptocurrency's price return and volume change for the upcoming days. Model performance is assessed by comparing predictions with actual observations using Mean Squared Error (MSE), which penalizes larger errors, and Mean Absolute Error (MAE), which provides a direct measure of average deviations.

\section{Experiment}
\label{sec4}
This section analyzes relationships between cryptocurrency markets and sentiment trends on Twitter and TikTok to answer \textbf{RQ1}. Integrating sentiment from both platforms improved forecast accuracy for Bitcoin and Dogecoin. We explore how social media signals reflect and potentially indicate price and volume changes.

\subsection{Cluster Analysis Results}
The cluster analysis of tweets was performed using K-means on TF-IDF data reduced to two principal components. 

\begin{figure}[htbp]
    \centering
    \begin{subfigure}[b]{0.48\linewidth}
        \centering
        \includegraphics[width=\linewidth]{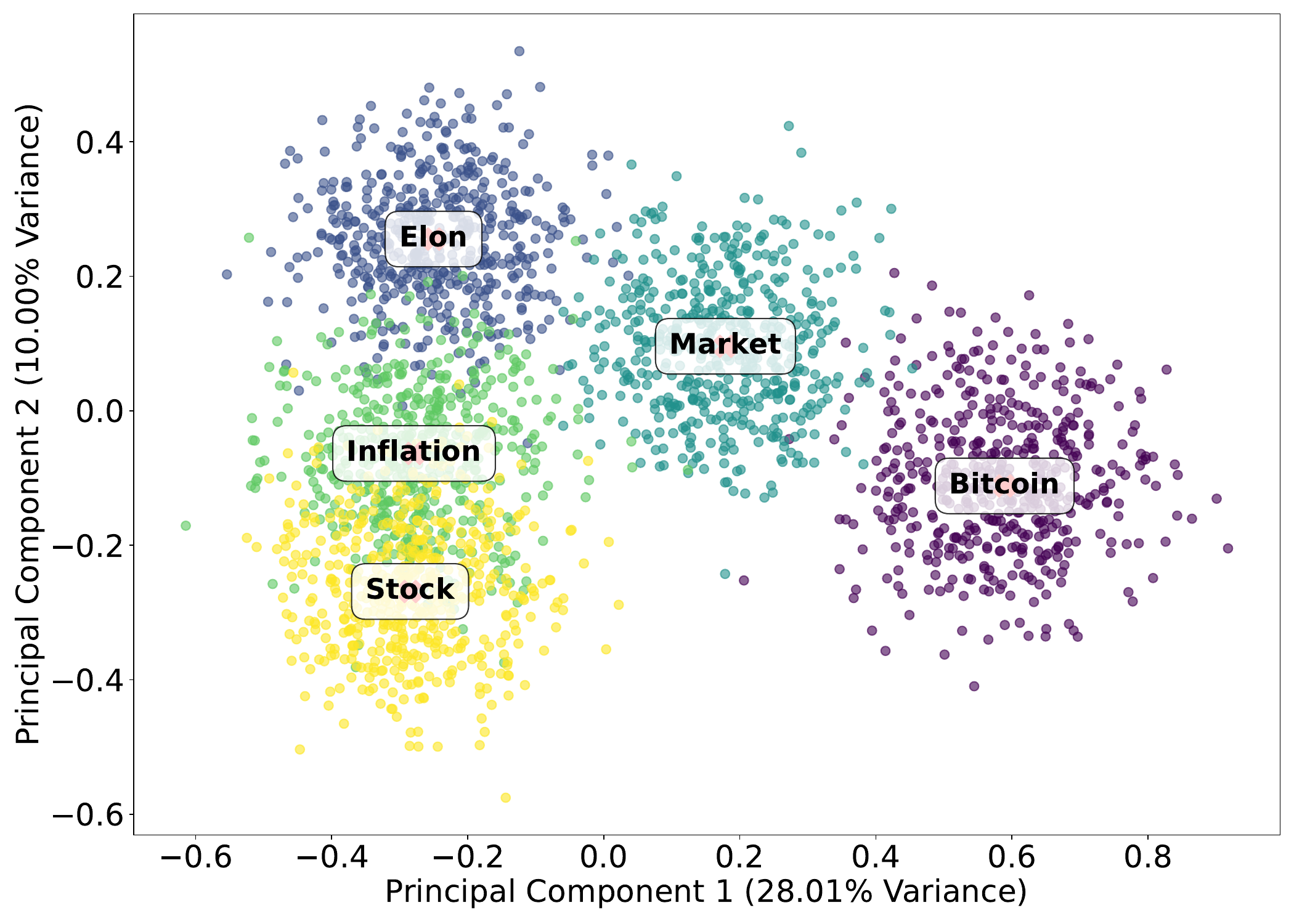}
        \caption{Twitter Cluster}
        \label{fig:cluster-twitter}
    \end{subfigure}
    \hfill
    \begin{subfigure}[b]{0.48\linewidth}
        \centering
        \includegraphics[width=\linewidth]{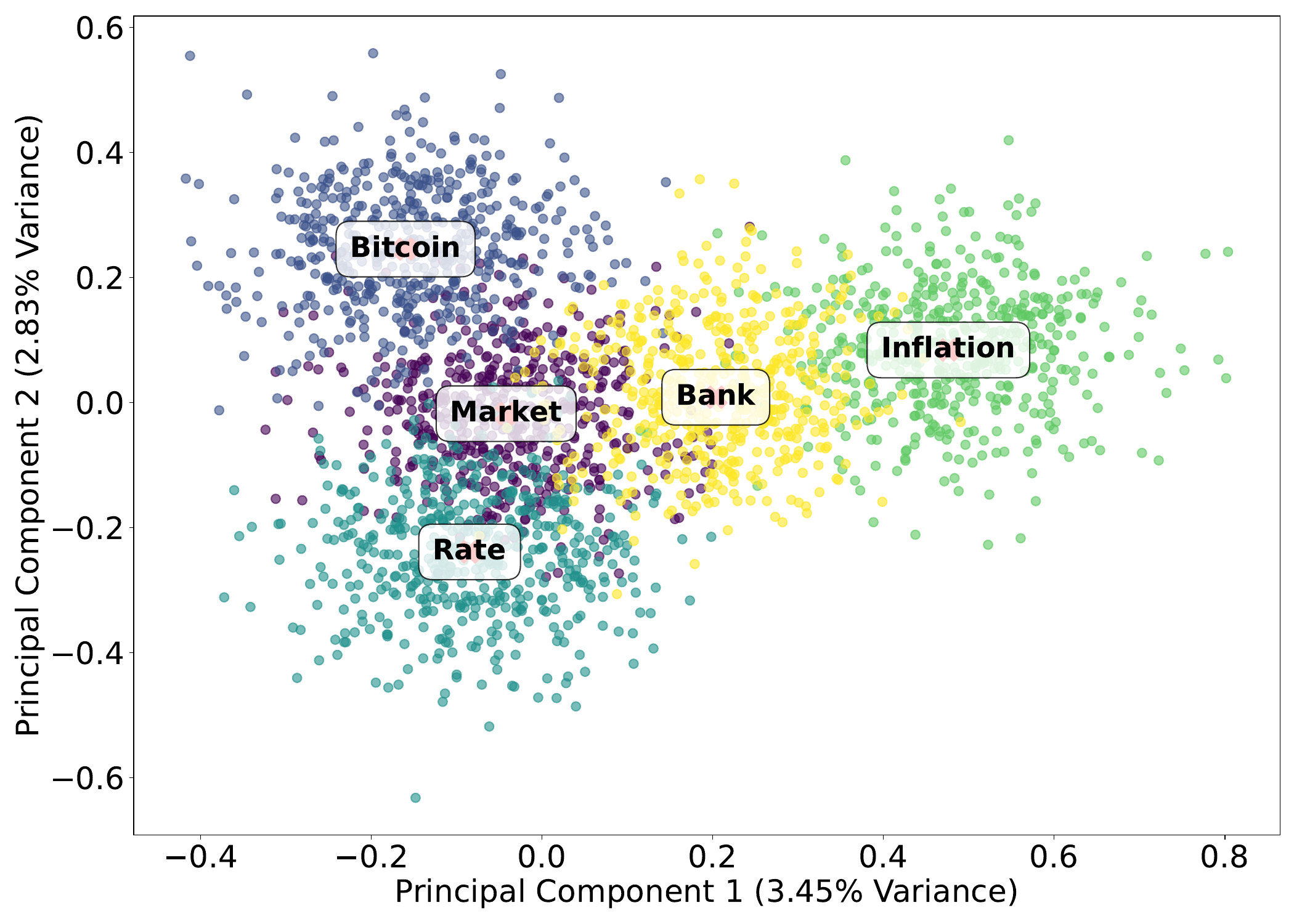}
        \caption{TikTok Cluster}
        \label{fig:cluster-tiktok}
    \end{subfigure}
    \caption{Cluster graphs based on the LDA results for Twitter and TikTok data.}
    \label{fig:cluster-lda-results}
\end{figure}

In Figure \ref{fig:cluster-lda-results} \subref{fig:cluster-twitter}, distinct themes and patterns emerge across the identified clusters. These patterns provide insights into how Twitter sentiment relates to cryptocurrency price returns and volume changes. The \emph{Bitcoin} and \emph{Stock} clusters are the largest, suggesting that discussions around Bitcoin and the stock market are highly influential in shaping sentiment. Tweets in the \emph{Bitcoin} cluster likely focus on cryptocurrency developments, Bitcoin price fluctuations, and news related to market events, which may directly correlate with price movements and volume changes in the cryptocurrency market. Similarly, the \emph{Stock} group offers a context that could also influence cryptocurrency sentiment, particularly when there is a global shock \cite{wang2020relationship, nguyen2022correlation}. The close proximity between the \emph{Stock}, \emph{Inflation}, and \emph{Elon} clusters points to the interconnectedness of stock market sentiment, economic conditions like inflation, and the influence of public figures such as Elon Musk. Musk’s statements often lead to shifts in cryptocurrency sentiment \cite{ante2023elon}. 

In Figure \ref{fig:cluster-lda-results} \subref{fig:cluster-tiktok} reveals five distinct clusters: \emph{Market}, \emph{Bitcoin}, \emph{Rate}, \emph{Inflation}, and \emph{Bank}. The \emph{Market} cluster aggregates discussions that pertain to overall market conditions, suggesting a broad focus on general economic and trading sentiments. The \emph{Bitcoin} cluster indicates a concentrated discourse on the flagship cryptocurrency, reflecting its prominence and influence. The \emph{Rate} and \emph{Inflation} clusters highlight topics related to macroeconomic factors, with \emph{Rate} focusing on interest rates and related financial policies, and \emph{Inflation} capturing concerns about rising prices and economic uncertainty. Finally, the \emph{Bank} cluster gathers content related to institutional finance, regulatory frameworks, and traditional banking operations.

\subsection{Rolling window coherence between crypto prices/volumes and sentiment}
This section examines the behavior of three selected cryptocurrencies BTC, DOGE, and USDT in relation to TikTok and Twitter sentiment across different time scales. BTC is included due to its status as the largest cryptocurrency by market capitalization \cite{meyer2023high,bohme2015bitcoin}. DOGE is chosen for its speculative nature and the unique influence of social media on its price movements, making it a prime example of sentiment-driven volatility \cite{ante2023elon,chohan2021history}. USDT, as a stablecoin, is analyzed for its relatively stable price behavior, offering insights into how sentiment interacts with low-volatility assets \cite{ante2021influence}.

\subsubsection{Price Return Coherence with Sentiment}

\begin{figure*}[htbp]
    \centering

    \begin{subfigure}[b]{0.48\textwidth}
        \centering
        \includegraphics[width=\linewidth]{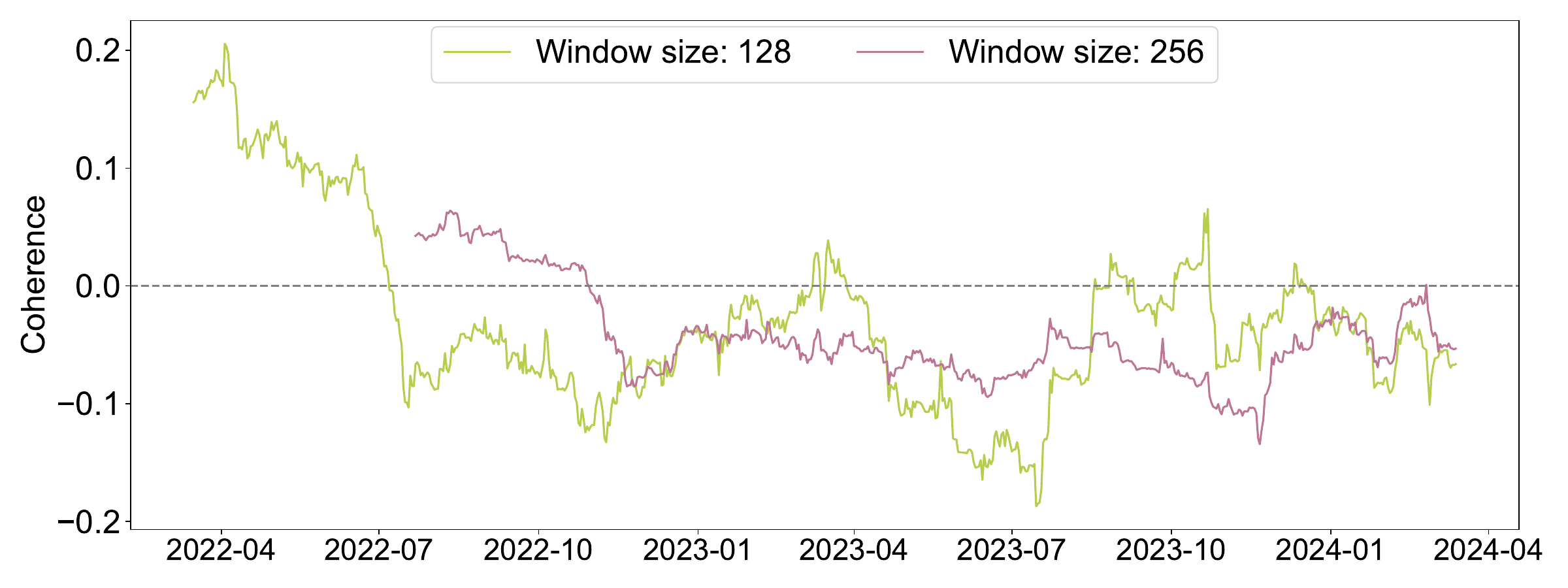}
        \caption{Bitcoin price returns and TikTok sentiment coherence}
        \label{fig:btc-prc-tiktok}
    \end{subfigure}
    \hfill
    \begin{subfigure}[b]{0.48\textwidth}
        \centering
        \includegraphics[width=\linewidth]{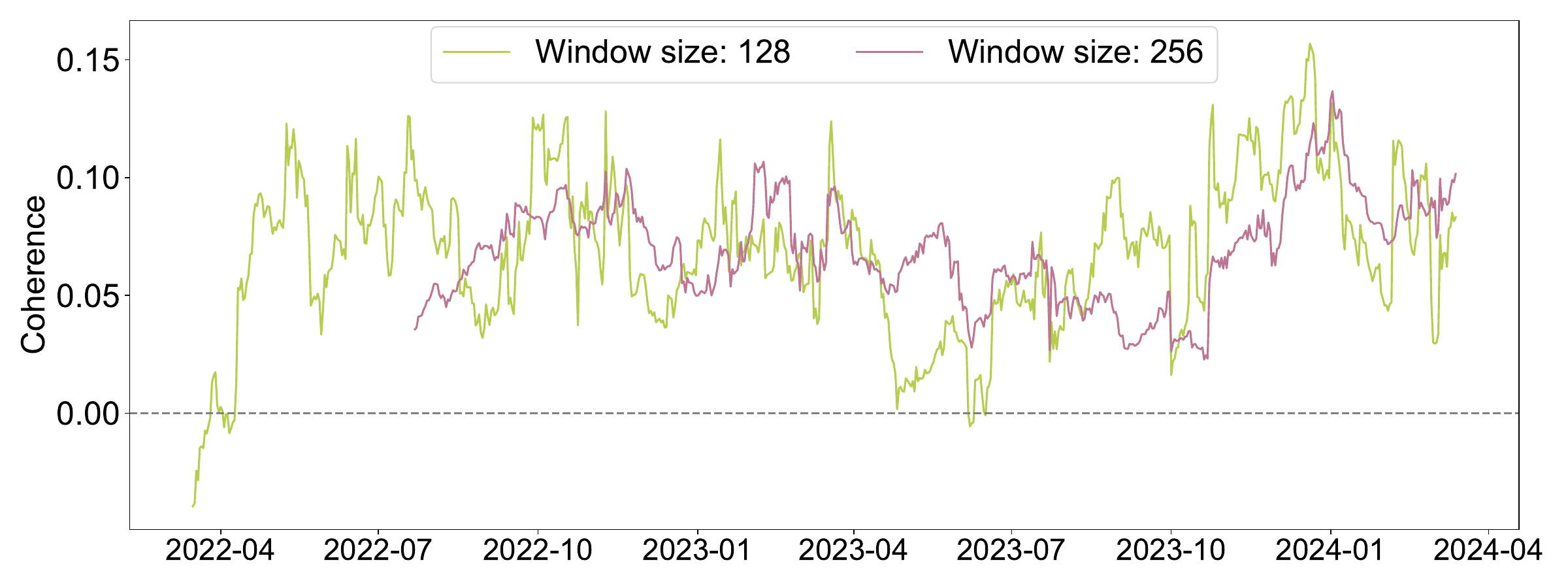}
        \caption{Bitcoin price returns and Twitter sentiment coherence}
        \label{fig:btc-prc-twitter}
    \end{subfigure}

    \vspace{0.5em} 

    \begin{subfigure}[b]{0.48\textwidth}
        \centering
        \includegraphics[width=\linewidth]{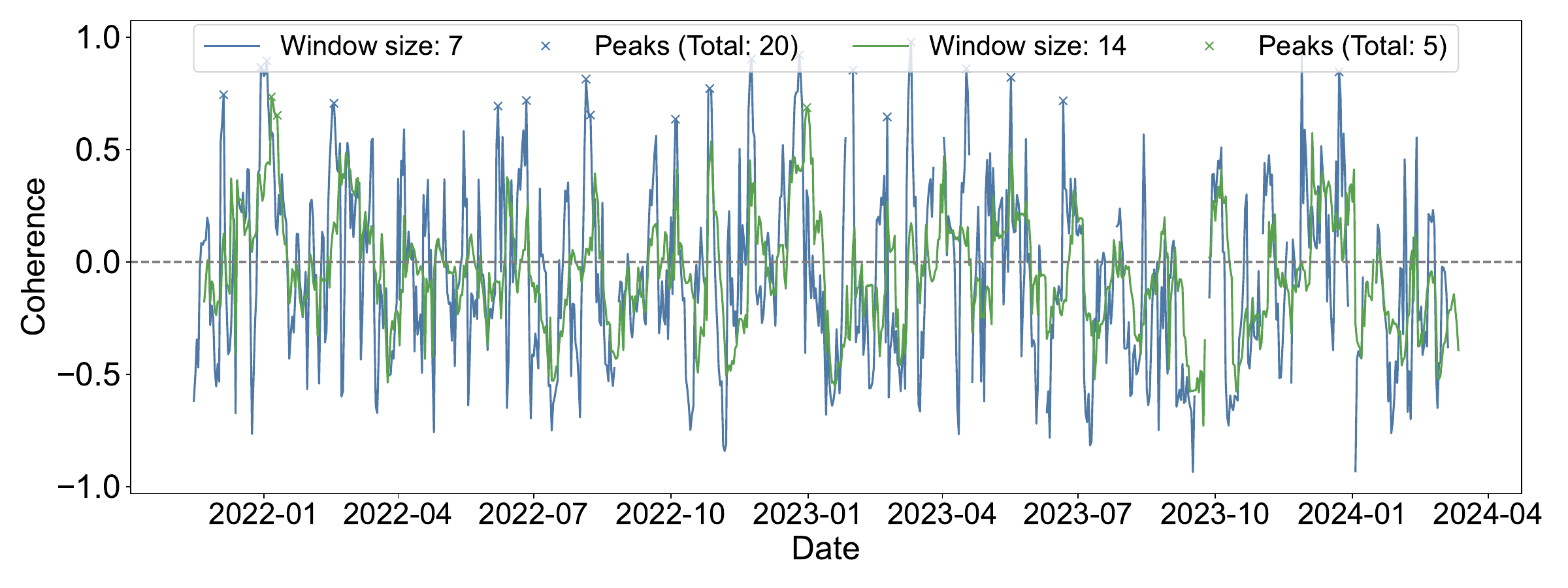}
        \caption{Dogecoin price returns and TikTok sentiment coherence}
        \label{fig:dog-prc-tiktok}
    \end{subfigure}
    \hfill
    \begin{subfigure}[b]{0.48\textwidth}
        \centering
        \includegraphics[width=\linewidth]{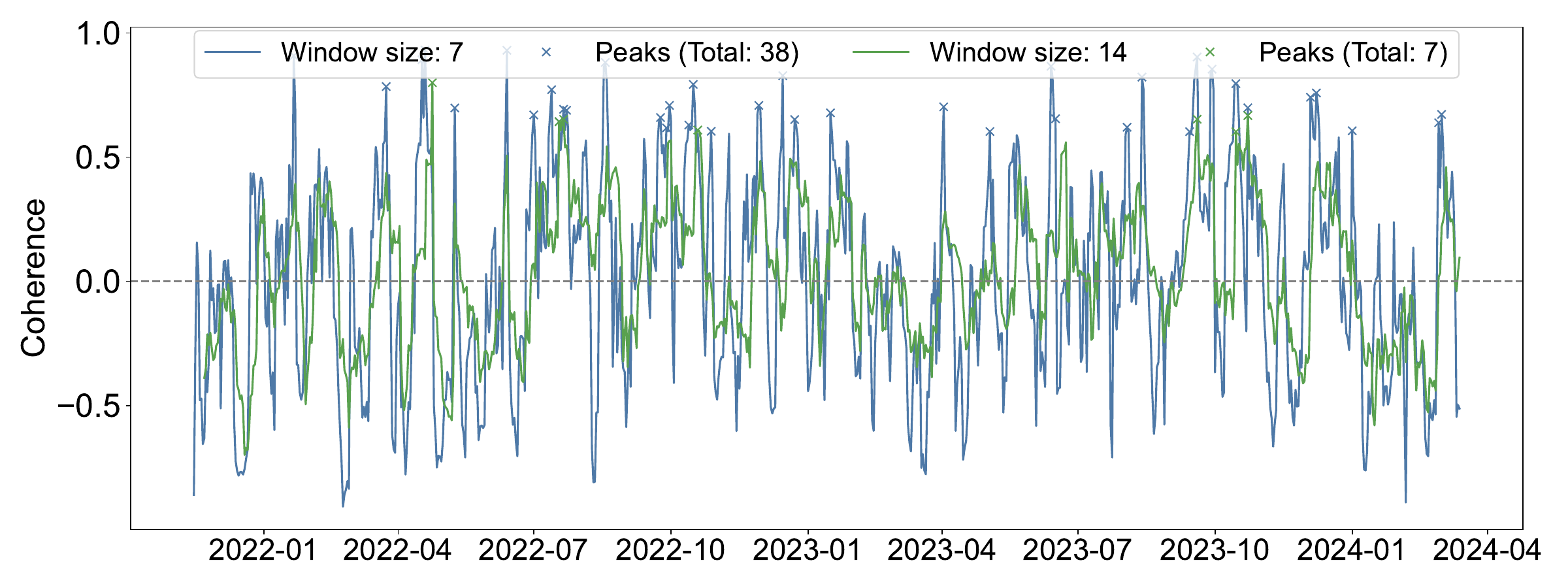}
        \caption{Dogecoin price returns and Twitter sentiment coherence}
        \label{fig:dog-prc-twitter}
    \end{subfigure}

    \vspace{0.5em} 

    \begin{subfigure}[b]{0.48\textwidth}
        \centering
        \includegraphics[width=\linewidth]{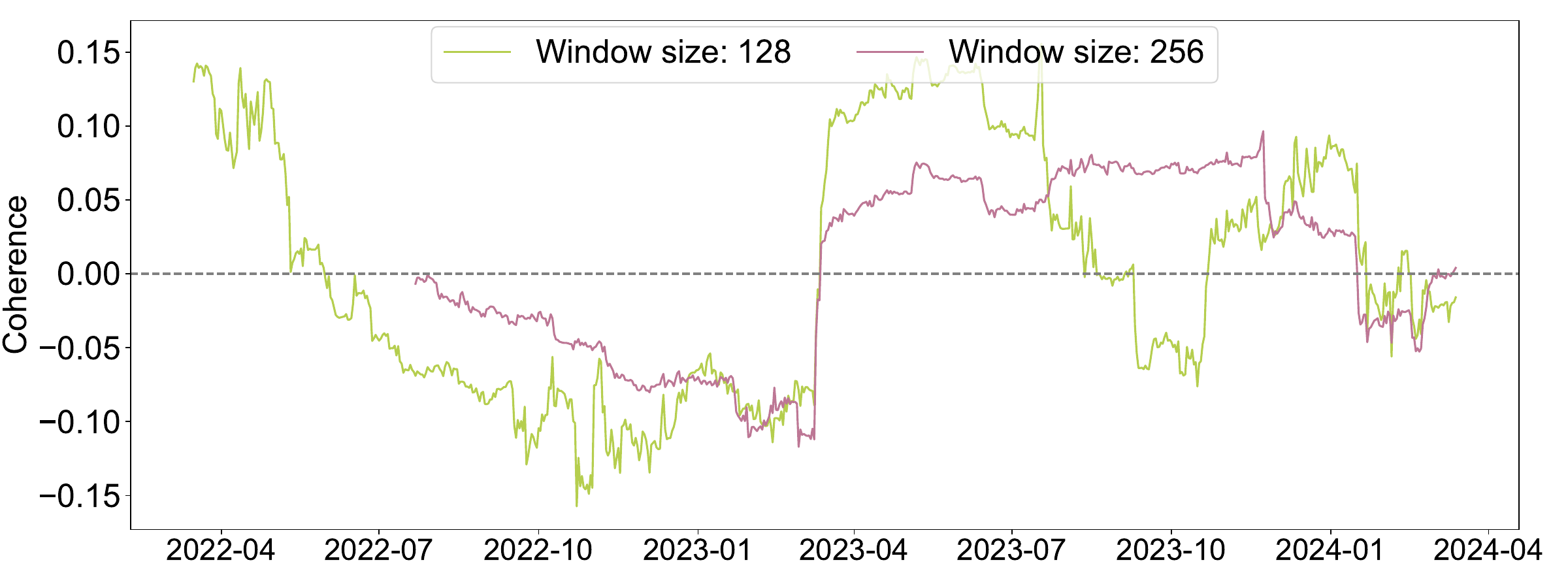}
        \caption{USDT price returns and TikTok sentiment coherence}
        \label{fig:usdt-prc-tiktok}
    \end{subfigure}
    \hfill
    \begin{subfigure}[b]{0.48\textwidth}
        \centering
        \includegraphics[width=\linewidth]{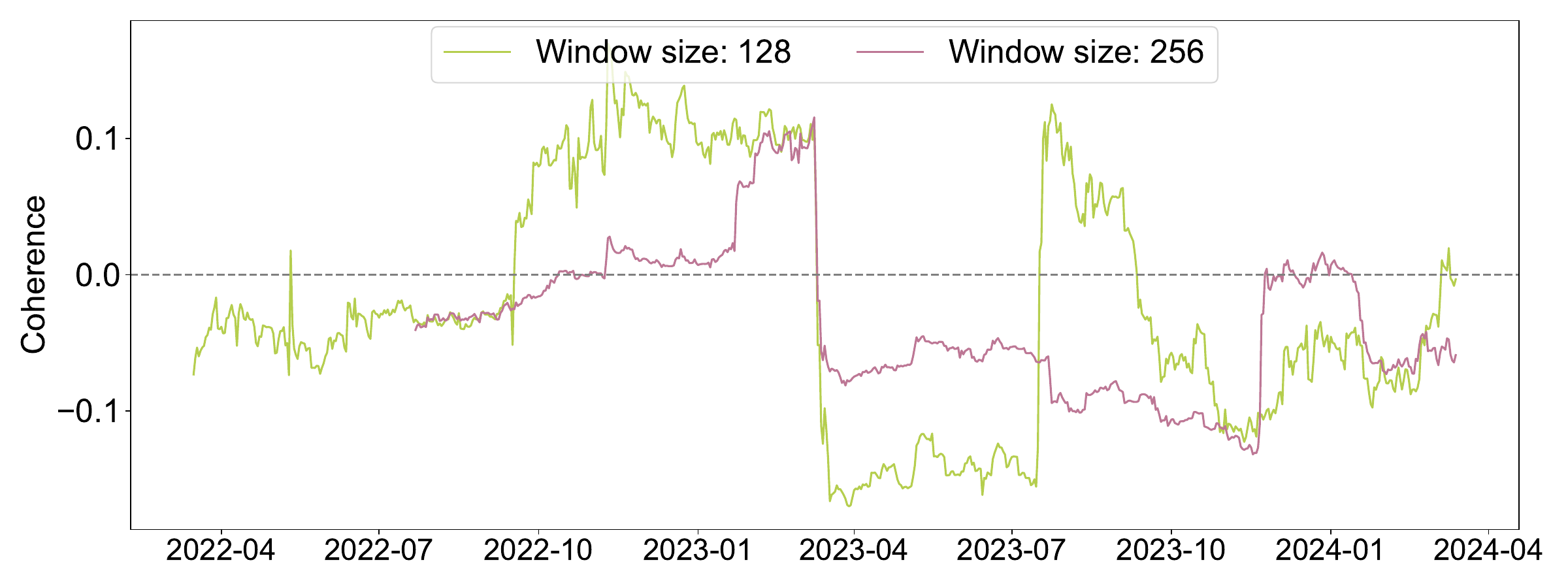}
        \caption{USDT price returns and Twitter sentiment coherence}
        \label{fig:usdt-prc-twitter}
    \end{subfigure}

    \caption{Rolling window coherence between cryptocurrency \textbf{price returns} and social media sentiment:
    (a,b) Bitcoin (long-term), (c,d) Dogecoin (short-term), (e,f) USDT (long-term).}
    \label{fig:crypto-rolling-window}
\end{figure*}

\textbf{BTC}: Figures \ref{fig:crypto-rolling-window}\subref{fig:btc-prc-tiktok} and \ref{fig:crypto-rolling-window}\subref{fig:btc-prc-twitter} represent the coherence between Bitcoin price returns and sentiment trends on TikTok and Twitter, respectively.

Figures \ref{fig:crypto-rolling-window}\subref{fig:btc-prc-tiktok} show a significant positive coherence between TikTok sentiment and Bitcoin returns in long-term windows from November 2021 to April 2022. This correlation suggests that positive TikTok sentiment is linked to rising Bitcoin returns during these periods, likely due to sustained public interest and the platform's younger user base \cite{montag2021psychology}. However, short-term coherence (7 and 14-day windows) is highly volatile, indicating that short-term TikTok sentiment fluctuations do not consistently align with immediate price movements. Overall, long-term coherence is mostly negative, with occasional positive spikes in January 2023, September 2023, and early 2024. This may result from content creators focusing on trending coins over Bitcoin to attract engagement, causing a disconnect between sentiment and price dynamics \cite{moser2023should}.

In contrast, Figures \ref{fig:crypto-rolling-window}\subref{fig:btc-prc-twitter} exhibits an initial negative coherence between Twitter sentiment and Bitcoin returns, suggesting a counter-cyclical relationship where negative sentiment coincides with high Bitcoin prices. This implies that Twitter users may react to elevated Bitcoin prices with skepticism or caution. Over time, long-term coherence becomes predominantly positive, indicating that Twitter sentiment aligns with Bitcoin trends over longer horizons \cite{kraaijeveld2020predictive, karalevicius2018using}.

The early positive coherence on TikTok can be partly attributed to Bitcoin prices remaining relatively high despite a decline after November 2021, maintaining public interest and optimism. TikTok's younger users, more open to new investment opportunities, likely boosted sentiment-driven engagement during this period \cite{montag2021psychology}.

Throughout the entire period, long-term coherence between TikTok sentiment and Bitcoin returns is largely negative, with few positive instances. This trend suggests that TikTok sentiment generally diverges from Bitcoin returns over extended timescales. One reason may be that TikTok creators prioritize trending coins with recent price surges over Bitcoin, leading to a mismatch between sentiment and price dynamics \cite{moser2023should}. Additionally, sharp declines in coherence after positive peaks indicate that temporary Bitcoin enthusiasm among TikTok’s younger audience fades quickly as prices fluctuate.

\textbf{DOGE}: Figures \ref{fig:crypto-rolling-window}\subref{fig:dog-prc-tiktok} and \ref{fig:crypto-rolling-window}\subref{fig:dog-prc-twitter} illustrate the coherence between Dogecoin price returns and sentiment trends on Tiktok and Twitter, respectively. 

In the short term, the coherence of TikTok and Twitter sentiment shows significant variability, with frequent shifts between positive and negative values. TikTok sentiment registers 25 peaks in total, which is fewer than the 45 peaks observed in Twitter sentiment. This suggests that Twitter sentiment exerts a more frequent and pronounced short-term effect on DOGE price returns, characterized by a highly reactive relationship often driven by temporary events such as responses to trending topics or market news. The greater number of peaks in Twitter sentiment coherence implies that Twitter users tend to respond quickly and intensely to short-term changes in the cryptocurrency market. TikTok emerged as a promotional platform for DOGE during the summer of 2020 \cite{nani2022doge}, while Musk’s tweets have been found to positively influence the returns and trading volume of DOGE \cite{ante2023elon}.

\textbf{USDT}: \ref{fig:crypto-rolling-window}\subref{fig:usdt-prc-tiktok} and \ref{fig:crypto-rolling-window}\subref{fig:usdt-prc-twitter} illustrate the coherence between USDT price returns and sentiment trends on Tiktok and Twitter, respectively. 

In the long term, coherence values for TikTok sentiment remain relatively stable and hover close to zero, with only minor positive and negative fluctuations. This indicates that TikTok sentiment does not exhibit a consistent or significant relationship with USDT price movements over extended periods. In contrast, coherence values for Twitter sentiment show a similar stability but with a noticeable upward shift beginning in mid-2022, suggesting a gradual increase in alignment between Twitter sentiment and USDT price returns. This upward trend could indicate that Twitter sentiment became more closely associated with long-term shifts in USDT prices. Both TikTok and Twitter coherence figures show a sharp change around April 2023, which may correspond to broader market or regulatory events affecting USDT. This observation aligns with findings from Osman et al. \cite{osman2024economic}, who noted that significant co-movement with sentiment and stablecoin.

\subsubsection{Volume Change Coherence with Sentiment}
\begin{figure*}[htbp]
    \centering

    \begin{subfigure}[b]{0.48\textwidth}
        \centering
        \includegraphics[width=\linewidth]{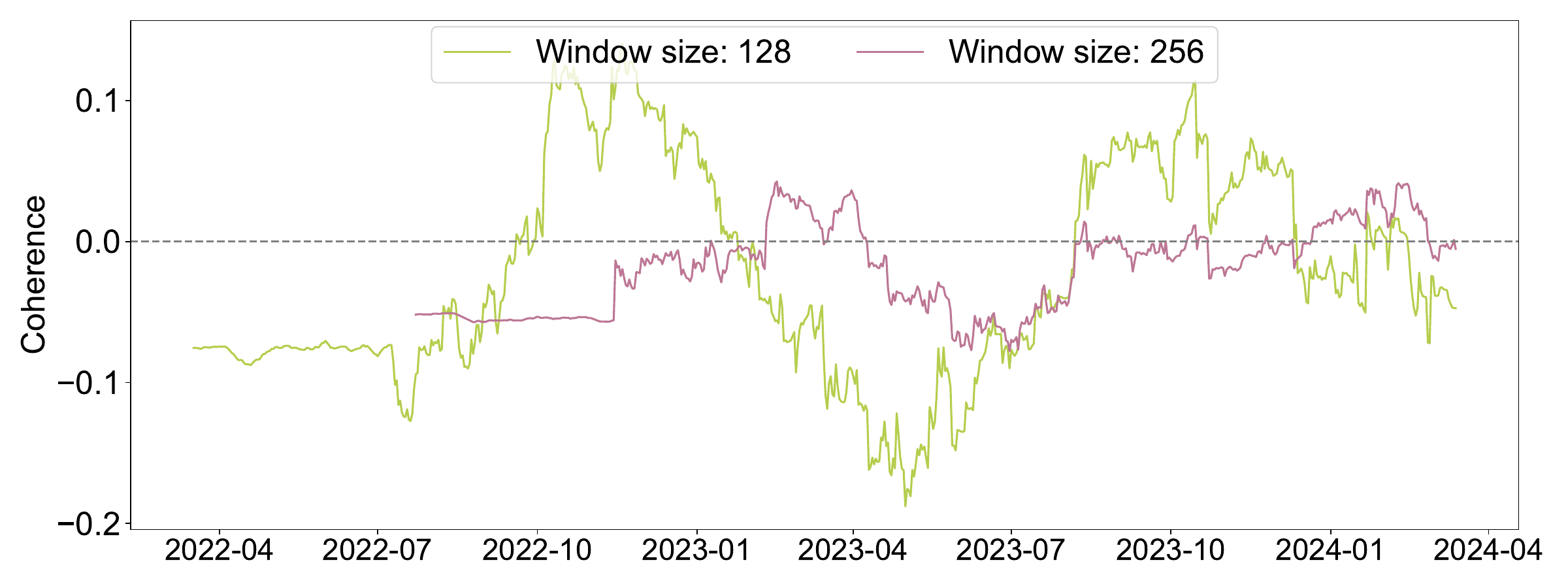}
        \caption{Bitcoin volume change and TikTok sentiment coherence}
        \label{fig:btc-vol-tiktok}
    \end{subfigure}
    \hfill
    \begin{subfigure}[b]{0.48\textwidth}
        \centering
        \includegraphics[width=\linewidth]{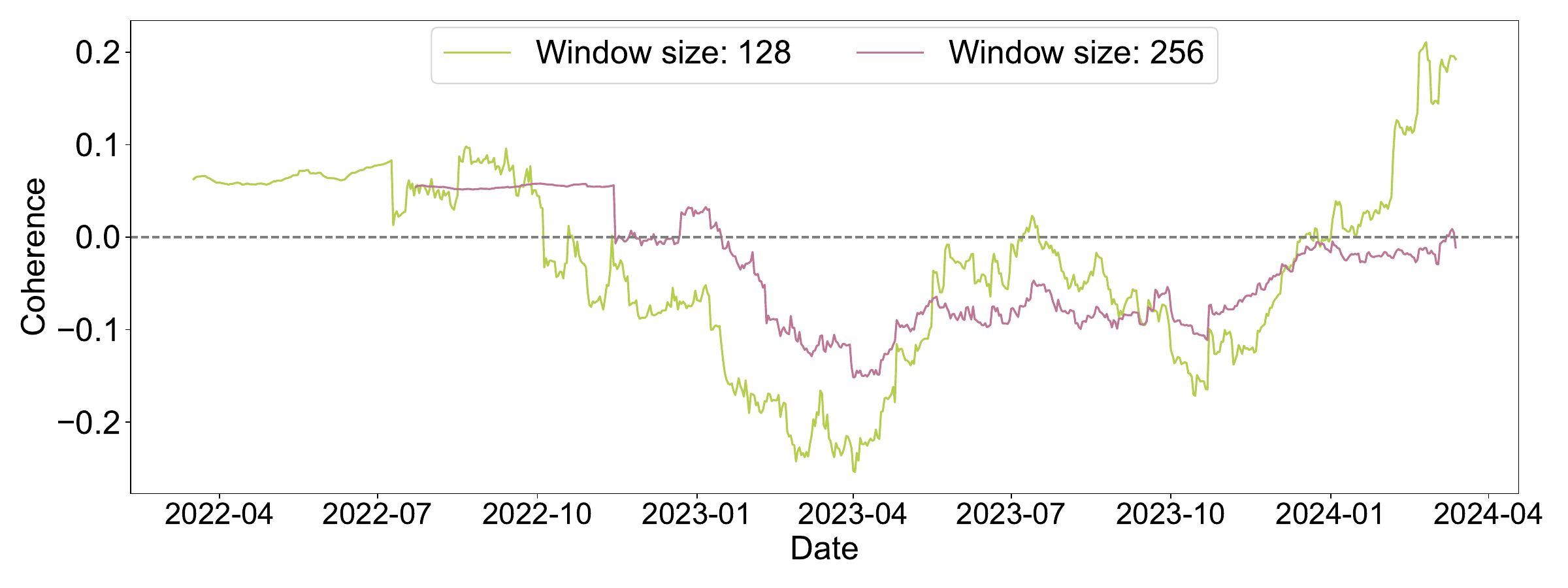}
        \caption{Bitcoin volume change and Twitter sentiment coherence}
        \label{fig:btc-vol-twitter}
    \end{subfigure}

    \vspace{0.5em} 

    \begin{subfigure}[b]{0.48\textwidth}
        \centering
        \includegraphics[width=\linewidth]{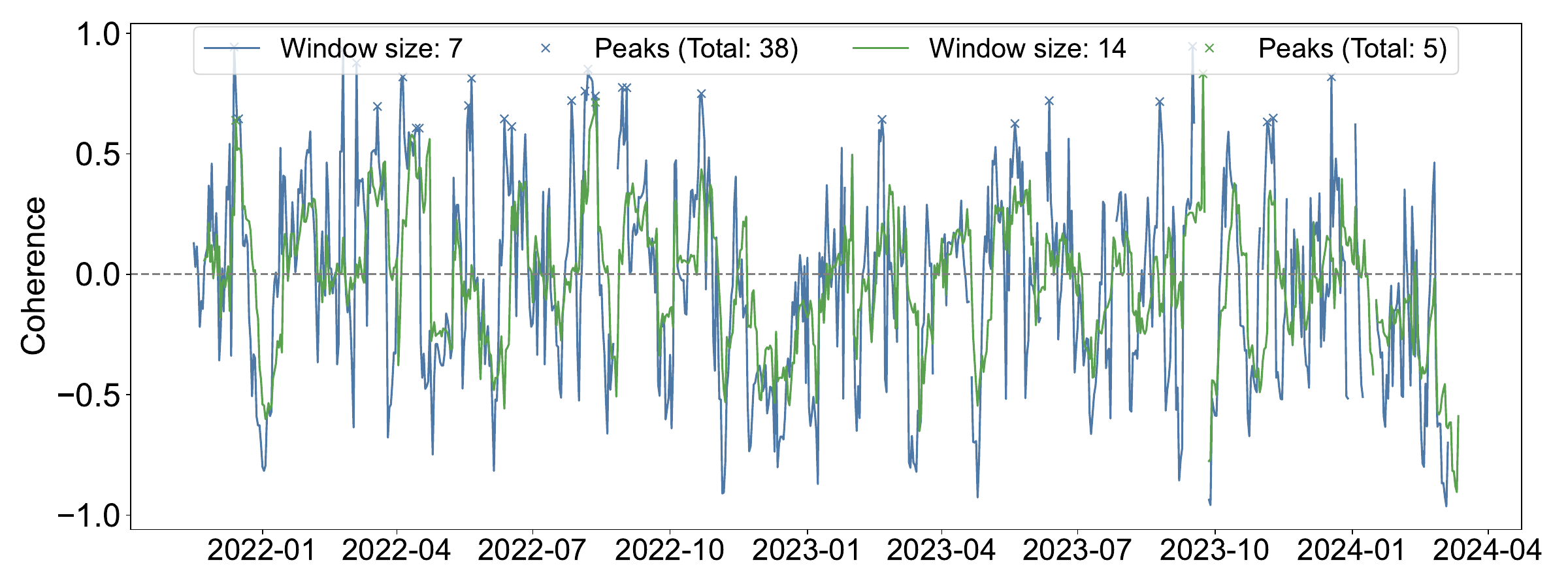}
        \caption{Dogecoin volume change and TikTok sentiment coherence}
        \label{fig:dog-vol-tiktok}
    \end{subfigure}
    \hfill
    \begin{subfigure}[b]{0.48\textwidth}
        \centering
        \includegraphics[width=\linewidth]{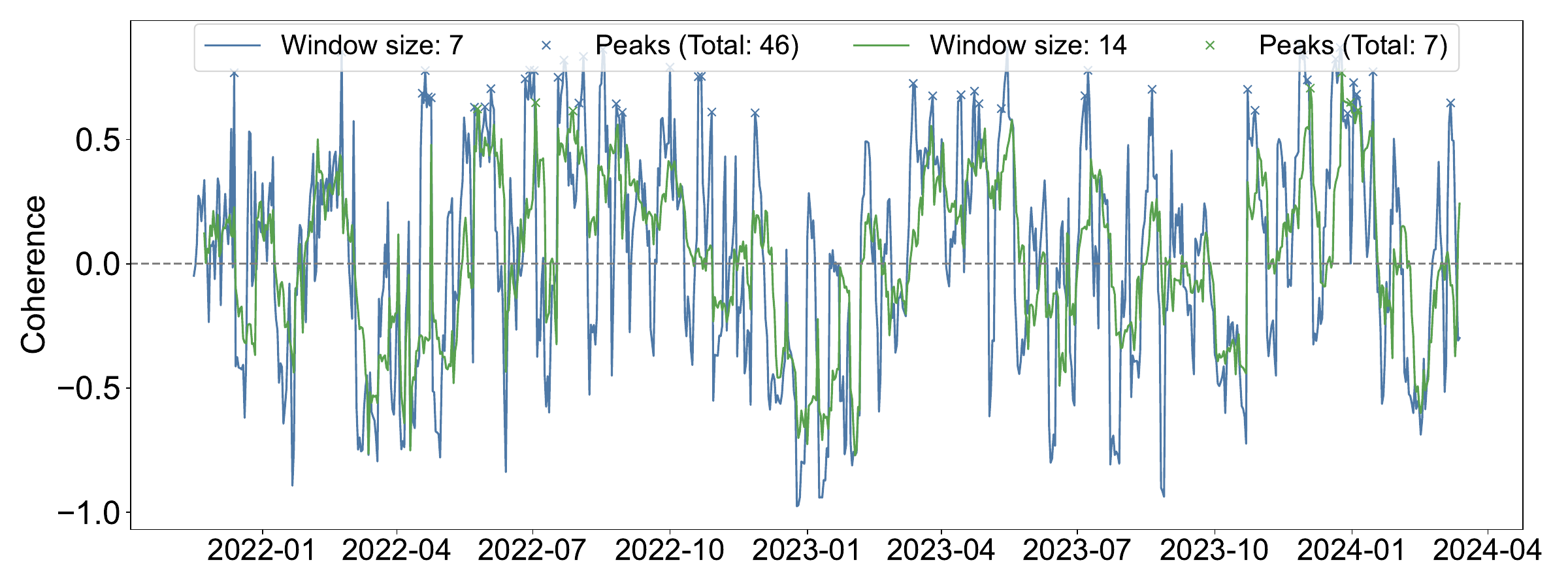}
        \caption{Dogecoin volume change and Twitter sentiment coherence}
        \label{fig:dog-vol-twitter}
    \end{subfigure}

    \vspace{0.5em} 

    \begin{subfigure}[b]{0.48\textwidth}
        \centering
        \includegraphics[width=\linewidth]{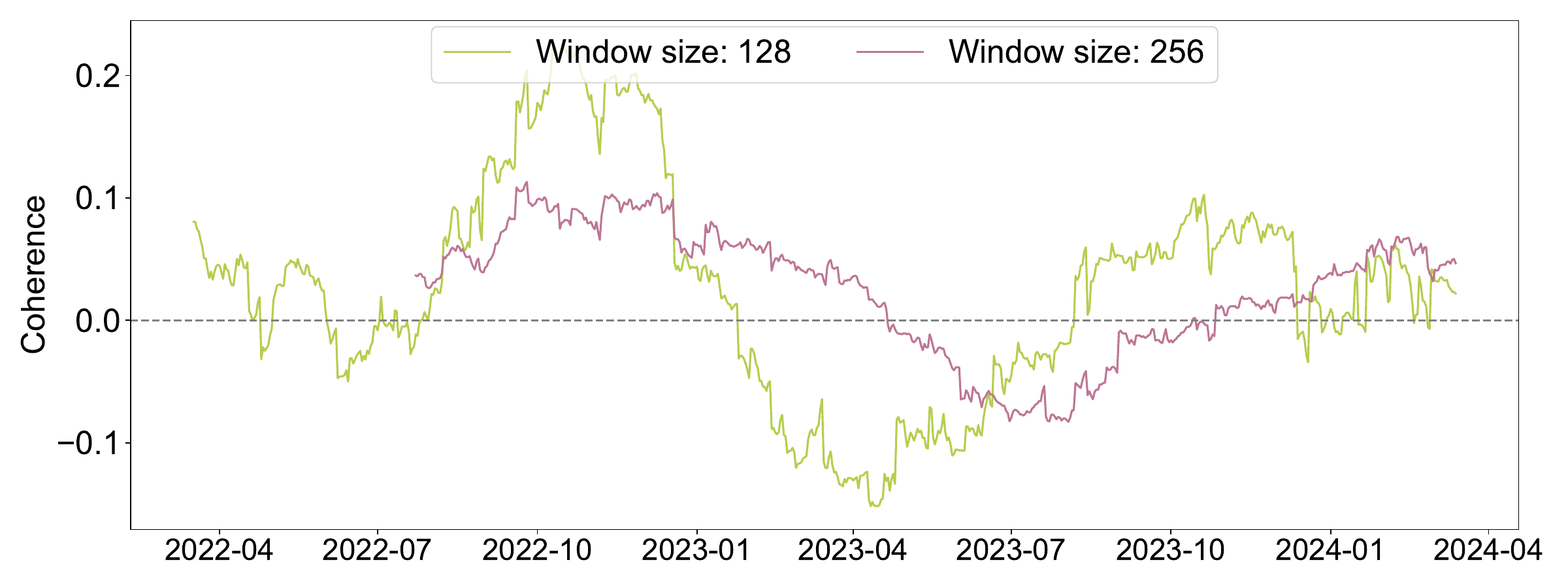}
        \caption{USDT volume change and TikTok sentiment coherence}
        \label{fig:usdt-vol-tiktok}
    \end{subfigure}
    \hfill
    \begin{subfigure}[b]{0.48\textwidth}
        \centering
        \includegraphics[width=\linewidth]{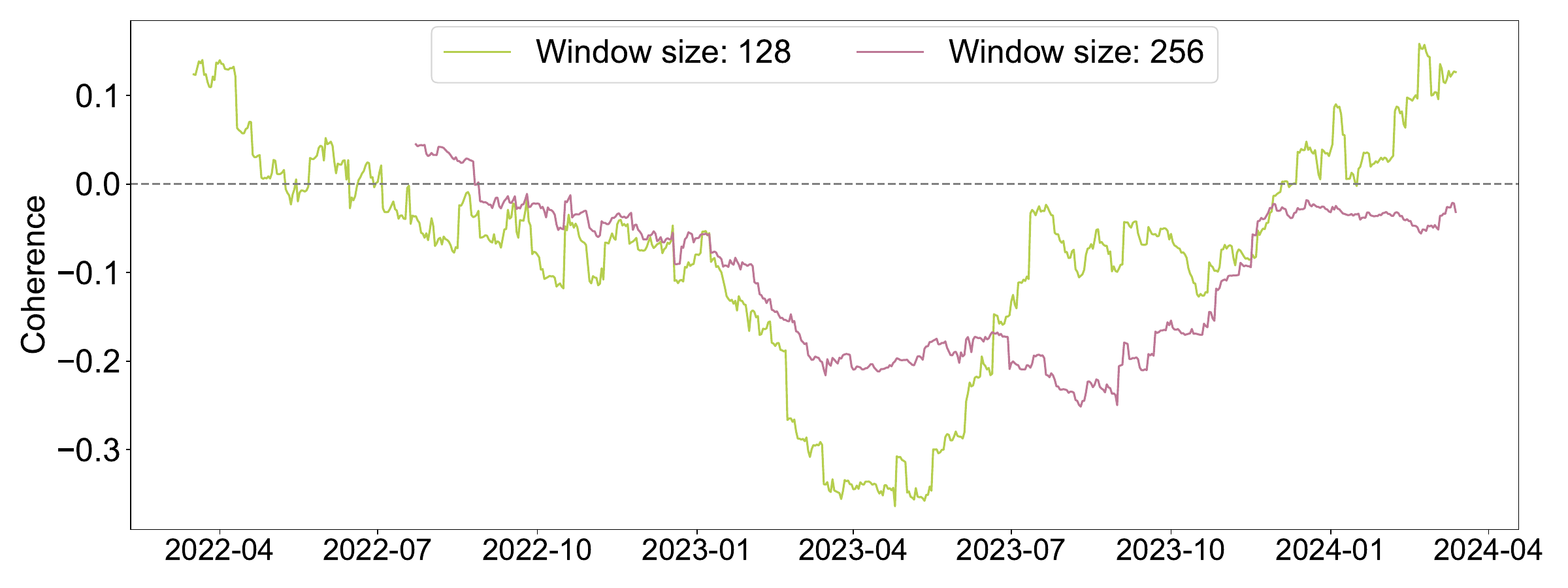}
        \caption{USDT volume change and Twitter sentiment coherence}
        \label{fig:usdt-vol-twitter}
    \end{subfigure}

    \caption{Rolling window coherence between cryptocurrency \textbf{volume changes} and social media sentiment:
    (a,b) Bitcoin (long-term), (c,d) Dogecoin (short-term), (e,f) USDT (long-term).}
    \label{fig:crypto-volume-rolling-window}
\end{figure*}

\textbf{BTC}: Figures \ref{fig:crypto-volume-rolling-window}\subref{fig:btc-vol-tiktok} and \ref{fig:crypto-volume-rolling-window}\subref{fig:btc-vol-twitter} illustrate the coherence between Bitcoin volume changes and sentiment trends on Tiktok and Twitter, respectively.

In Fig.\ref{fig:crypto-volume-rolling-window}\subref{fig:btc-vol-tiktok}, the coherence between TikTok sentiment and Bitcoin volume changes is relatively weak overall. Over the long term, coherence values are predominantly centered around zero, indicating minimal alignment between TikTok sentiment and BTC trading activity. However, notable periods of stronger positive and negative coherence are observed, particularly during early 2023 and late 2023, when changes in TikTok sentiment aligned more closely with BTC volume fluctuations. These temporary alignments may reflect heightened social media attention or specific events that captured user interest in Bitcoin. Despite these sporadic peaks, most coherence values are negative, suggesting that TikTok sentiment generally lacks a sustained or consistent influence on BTC trading volume over extended periods. Instead, its impact appears limited to brief bursts of alignment during high-interest periods, likely driven by trend-focused content creation on the platform.

Similarly, in Fig.\ref{fig:crypto-volume-rolling-window}\subref{fig:btc-vol-twitter}, coherence between Twitter sentiment and BTC volume changes also remains weak in the long term, with values largely stable and near zero. Slight positive intervals are observed, most notably in early 2024, indicating occasional alignment between Twitter sentiment and BTC trading activity. However, these instances are minimal and inconsistent, underscoring Twitter’s limited influence on long-term volume trends. Mai et al. \cite{mai2015impacts} reported similar findings, noting a weak correlation between social media activity and Bitcoin trading volume, particularly during periods of heightened engagement.

\textbf{DOG}: Figures \ref{fig:crypto-volume-rolling-window}\subref{fig:dog-vol-tiktok} and \ref{fig:crypto-volume-rolling-window}\subref{fig:dog-vol-twitter} 
illustrate the coherence between Dogecoin volume changes and sentiment trends on Tiktok and Twitter, respectively.

In the short term, 43 peaks are observed in TikTok sentiment coherence, which is fewer than the 53 peaks found in Twitter sentiment coherence. This suggests that Twitter sentiment may exert a more frequent and reactive impact on DOGE trading volume, likely due to its function as a platform for real-time news and market conversations. The higher number of coherence peaks in Twitter sentiment reflects the tendency of Twitter users to respond quickly and intensely to short-term market changes. Ante \cite{ante2023elon} found that tweets related to DOGE consistently lead to significant positive returns and increased trading volume. This evidence supports the idea that Twitter sentiment can amplify short-term trading activity in the cryptocurrency market by rapidly mirroring shifts in trader sentiment and market focus.

\textbf{USDT}: Figures \ref{fig:crypto-volume-rolling-window}\subref{fig:usdt-vol-tiktok} and \ref{fig:crypto-volume-rolling-window}\subref{fig:usdt-vol-twitter} illustrate the coherence between USDT volume changes and sentiment trends on Tiktok and Twitter, respectively. 

In the long term, TikTok sentiment coherence exhibits a slight upward shift beginning in mid-2022, suggesting a closer alignment with USDT trading volume. This trend may reflect increased social media engagement surrounding stablecoins. Au et al. \cite{au2024characteristics} highlight that the stability and widespread adoption of stablecoins like Tether are heavily influenced by public trust, reserve backing, and their ability to maintain price stability. Conversely, Twitter sentiment coherence remains negative during this period, indicating a misalignment with USDT trading activity.

\subsection{Dynamic connectedness among sentiment and cryptocurrency prices and volumes}
\subsubsection{Unit root test}
The unit root test results presented in Table \ref{tab:unit_root}, performed using the Augmented Dickey-Fuller (ADF) technique, reveal statistically significant coefficients for all variables at the 1\% significance level, suggesting that all-time series are stationary at the first difference. These preliminary results justify the use of DCC-GARCH-based connectedness models in the study, as it requires strict stationarity.

\begin{table}[ht]
\centering
\caption{Results of Unit Root Tests for Price, Volume, and Sentiment}
\label{tab:unit_root}
\resizebox{\columnwidth}{!}{%
\begin{tabular}{lcccccccccccc}
\toprule
\textbf{Test} 
& \textbf{BTCPRC} & \textbf{ETHPRC} & \textbf{SOLPRC} & \textbf{DOGPRC} & \textbf{XRPPRC} & \textbf{USDCPRC} & \textbf{USDTPRC} \\
\midrule
ADF Statistics 
& 638.16*** & 541.12*** & 1476.42*** & 6694.37*** & 219593.15*** & 4560786.4*** & 50821.13*** \\
Jarque--Bera    
& -9.73** & -9.42** & -9.32** & -9.95** & -9.60** & -14.41** & -11.85** \\
\midrule
\textbf{Test} 
& \textbf{BTCVOL} & \textbf{ETHVOL} & \textbf{SOLVOL} & \textbf{DOGVOL} & \textbf{XRPVOL} & \textbf{USDCVOL} & \textbf{USDTVOL} \\
\midrule
ADF Statistics 
& 7099.2*** & 172.26*** & 1015519.61*** & 98.6*** & 309.11*** & 22584.28*** & 517.65*** \\
Jarque--Bera    
& -9.74** & -13.67** & -13.12** & -15.35** & -14.77** & -13.66** & -12.28** \\
\midrule
\textbf{Test} 
& \textbf{Twitter Sent.} & \textbf{TikTok Sent.} \\
\midrule
ADF Statistics 
& 129.04*** & 15.18*** \\
Jarque--Bera    
& -6.83** & -8.08** \\
\bottomrule
\end{tabular}%
}
\end{table}

\subsubsection{Dynamic Connectedness and Pairwise Network Analysis}
\begin{figure}[htbp]
    \centering
    \begin{subfigure}[b]{0.48\linewidth}
        \centering
        \includegraphics[width=\linewidth]{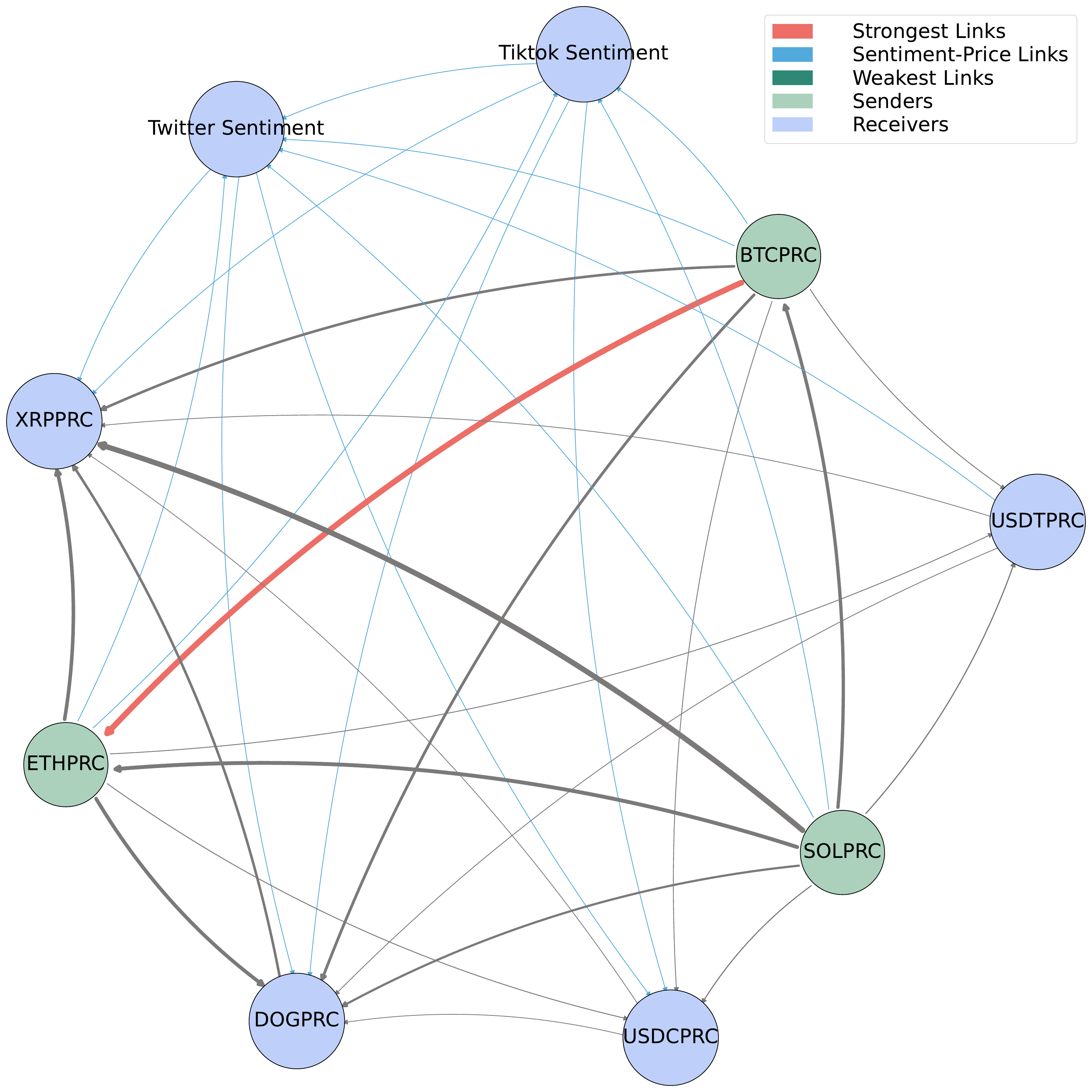}
        \caption{Connectedness among sentiment and prices}
        \label{fig:price-connect}
    \end{subfigure}
    \hfill
    \begin{subfigure}[b]{0.48\linewidth}
        \centering
        \includegraphics[width=\linewidth]{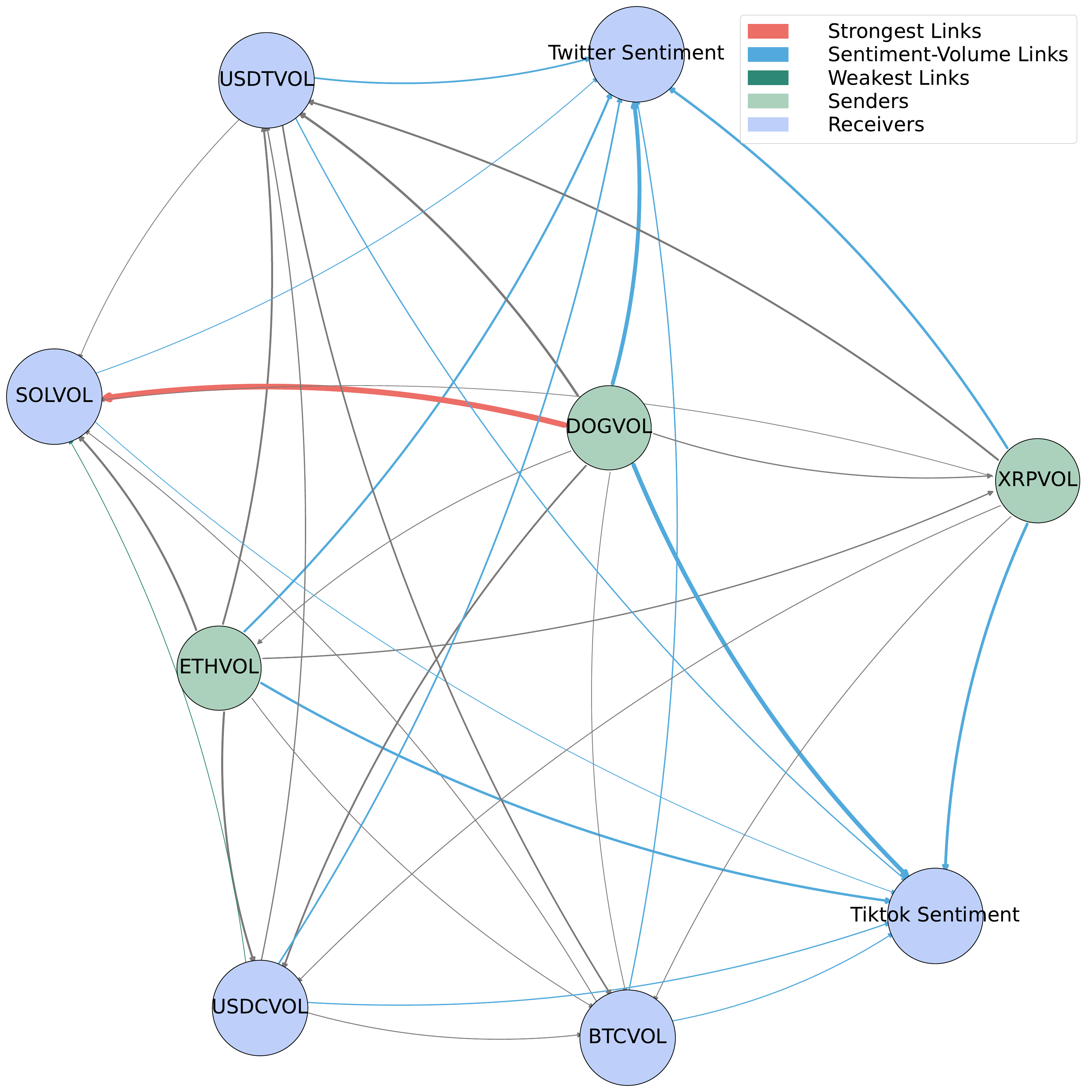}
        \caption{Connectedness among sentiment and volumes}
        \label{fig:volume-connect}
    \end{subfigure}

    \caption{Network graphs for connectedness among sentiment, prices, and volumes.}
    \label{fig:connect-overall}
\end{figure}

The network graphs in Figs. \ref{fig:price-connect} and \ref{fig:volume-connect} further illustrate these dynamics. BTC, ETH, and SOL are the strongest transmitters of price return shocks, while XRP and DOGE act as net receivers. For volume changes, DOGE is the primary transmitter, influencing SOL and sentiment indices. Sentiment indices show limited outward influence but are highly sensitive to external shocks, consistent with Ji et al. \cite{ji2019dynamic}. The ETH-to-XRP connection observed in the network graph suggests evolving market relationships, aligning with Koutmos \cite{koutmos2018return}, who highlighted strong interdependencies among cryptocurrencies.

Smales \cite{smales2022investor} emphasizes that large-market-cap cryptocurrencies like BTC and ETH serve as conduits for risk transmission, influencing smaller assets through market reactions and media coverage. DOGE, despite its lower market capitalization, exhibits dual roles: it is a significant receiver of price return shocks and a primary transmitter of volume changes. These findings reinforce DOGE’s speculative nature and its unique position within the market.

Our results partially align with Li et al. \cite{li2020risk}, who noted that smaller-market-cap cryptocurrencies often transmit spillovers to larger ones. XRP follows this pattern, while DOGE deviates, emerging as the largest receiver of price return shocks but a key transmitter of volume changes. Fig. \ref{fig:volume-connect} illustrates DOGE’s impact on SOL volume and sentiment indices, while ETH volume also acts as a key transmitter to TikTok sentiment. These findings align with Balcilar et al. \cite{balcilar2017can}, who emphasized the critical role of volume in shaping market dynamics. The results highlight the interconnected nature of cryptocurrency markets and the importance of considering both price and volume dynamics in understanding market relationships.

\section{Forecasting}

Given TikTok’s influence on altcoins and Twitter’s broader impact on the cryptocurrency market, this section introduces a prediction task to assess the predictive power of sentiment across different time scales to answer \textbf{RQ2}. Through this analysis, we aim to determine how sentiment data from these platforms can enhance forecasting accuracy and provide insights into sentiment’s role in driving market trends.

\subsubsection{Time-Series Forecasting with Transformer-based Model}
This section outlines the methodology utilized to predict cryptocurrency price returns and volume changes using the \texttt{MOMENT} model, as described by Goswami et al. \cite{goswami2024moment}. The \texttt{MOMENT} model employs transformer models pre-trained on a masked time series prediction task, incorporating innovations such as masked representation learning and cross-modal transfer learning. The model excels in various time-series tasks, including classification, anomaly detection, imputation, and long-horizon forecasting, outperforming the latest models like Time-LLM \cite{jin2023time} and GPT4TS \cite{zhou2023one} as well as ARIMA.

\subsubsection{Experiment Setup}
The target variables are the daily returns and volume changes for each specific cryptocurrency. The data is partitioned into training, validation, and testing sets, with 60\% allocated for training, 10\% for validation, and 30\% for testing, following the methodology recommended by Xu and Goodacre \cite{xu2018splitting}. Additionally, we employ the wavelet method to decompose the cryptocurrency data into multiple time scales: 7, 14, 32, 64, 128, and 256. Each forecasting scenario considered Twitter sentiment, TikTok sentiment, and a combination of both Twitter and TikTok sentiment. We trained model each 10 epochs for 100 epochs in sum.

\subsubsection{Results}
Here, we use $MOMENT_{Twitter}$ as our baseline, given extensive research leveraging Twitter sentiment for cryptocurrency forecasting \cite{kraaijeveld2020predictive, saleem2024twitter, wolk2020advanced, aslam2022sentiment}. In contrast, TikTok sentiment remains underexplored, making Twitter sentiment a suitable benchmark for assessing TikTok's performance. The model was trained separately with each sentiment dataset, and the fit distributions for each metric were compared to the baseline. Comparative results are presented in Table \ref{tab:forecasting}.

In forecasting Bitcoin price returns (BTCPRC), the combined sentiment model ($MOMENT_{Twitter,TikTok}$) consistently yields lower MSE and MAE across short-, medium-, and long-term horizons. Notably, in the long term, the combined model outperforms the Twitter-only model by roughly 20\%, demonstrating that integrating diverse sentiment sources enhances predictive accuracy. For Bitcoin trading volume (BTCVOL), the combined model again produces the lowest error values, capturing trading dynamics more effectively \cite{zhou2023multi}.

For DOGPRC, $MOMENT_{TikTok}$ achieves the lowest errors in the short term, outperforming $MOMENT_{Twitter}$ by about 35\%. In the medium term, Twitter sentiment aligns more closely with Dogecoin's price movements. Over the long term, the combined model delivers superior performance, reflecting DOGE’s speculative nature and providing a stronger signal for forecasting DOGE trading volume \cite{li2024sentiment, osman2024economic}.

For the stablecoin group, none of the sentiment indicators show significant forecasting improvements, likely due to their inherent stability and low volatility. As these assets are typically pegged to traditional currencies, market fluctuations are minimal, reducing the predictive relevance of sentiment-based models \cite{lket2023drives}. This result underscores that while sentiment signals are impactful for volatile cryptocurrencies, their influence on stablecoins is limited.

\begin{table*}[htbp]
\centering
\caption{Forecasting with Twitter and TikTok Sentiment - Price and Volume}
\resizebox{0.75\linewidth}{!}{
\begin{tabular}{cl|ccccc|ccccc}
\toprule
\textbf{} & \textbf{} & \multicolumn{5}{c}{\textbf{Price}} & \multicolumn{5}{c}{\textbf{Volume}} \\
\cmidrule(lr){3-7} \cmidrule(lr){8-12}
\textbf{} & \textbf{} & \multicolumn{2}{c}{\textbf{BTCPRC}} & \multicolumn{2}{c}{\textbf{DOGPRC}} & & \multicolumn{2}{c}{\textbf{BTCVOL}} & \multicolumn{2}{c}{\textbf{DOGVOL}} & \\
\cmidrule(lr){3-4} \cmidrule(lr){5-6} \cmidrule(lr){8-9} \cmidrule(lr){10-11}
\textbf{Window Size} & \textbf{Metric} & \textbf{MSE} & \textbf{MAE} & \textbf{MSE} & \textbf{MAE} & & \textbf{MSE} & \textbf{MAE} & \textbf{MSE} & \textbf{MAE} \\
\midrule

\multirow{3}{*}{ Short-term (7)} & \textbf{$MOMENT_{Twitter}$} & 1.373 & 0.792 & 1.063 & 0.721 & & 1.301 & 0.779 & 0.864 & 0.649 \\
 & \textbf{$MOMENT_{TikTok}$} & 1.280 & 0.695 & \textbf{0.682} & \textbf{0.647}  & & 1.387 & 0.822 & 0.943 & 0.689 \\
 & \textbf{$MOMENT_{Twitter,TikTok}$} & \textbf{1.126} & \textbf{0.619} & 0.870 & 0.697 & & \textbf{1.081} & \textbf{0.713} & \textbf{0.787} & \textbf{0.625} \\
\midrule

\multirow{3}{*}{Short-term (14)} & \textbf{$MOMENT_{Twitter}$} & 1.380 & 0.794 & 1.078 & 0.725 & & 1.310 & 0.782 & 0.880 & 0.655 \\
 & \textbf{$MOMENT_{TikTok}$} & 1.285 & 0.698 & \textbf{0.885} & \textbf{0.653} & & 1.394 & 0.824 & 0.955 & 0.694 \\
 & \textbf{$MOMENT_{Twitter,TikTok}$} & \textbf{1.135} & \textbf{0.623} & 1.008 & 0.688 & & \textbf{1.092} & \textbf{0.717} & \textbf{0.801} & \textbf{0.632} \\
\midrule

\multirow{3}{*}{Medium-term (32)} & \textbf{$MOMENT_{Twitter}$} & 1.303 & 0.709 & \textbf{0.913} & \textbf{0.661} & & 1.337 & 0.791 & 0.916 & 0.667 \\
 & \textbf{$MOMENT_{TikTok}$} & 1.310 & 0.705 & 1.102 & 0.699 & & 1.404 & 0.825 & 0.973 & 0.698 \\
 & \textbf{$MOMENT_{Twitter,TikTok}$} & \textbf{1.153} & \textbf{0.628} & 1.047 & 0.728 & & \textbf{1.111} & \textbf{0.725} & 0.956 & 0.641 \\
\midrule

\multirow{3}{*}{ Medium-term (64)} & \textbf{$MOMENT_{Twitter}$} & 1.286 & 0.705 &  \textbf{0.902} & \textbf{0.652} & & 1.334 & 0.805 & 0.920 & 0.675 \\
 & \textbf{$MOMENT_{TikTok}$} & 1.395 & 0.810 & 1.069 & 0.704 & & 1.350 & 0.799 & 0.929 & 0.671 \\
 & \textbf{$MOMENT_{Twitter,TikTok}$} & \textbf{1.122} & \textbf{0.620} & 1.060 & 0.703 & & \textbf{1.087} & \textbf{0.717} & 0.898 & 0.632 \\
\midrule

\multirow{3}{*}{ Long-term (128)} & \textbf{$MOMENT_{Twitter}$} & 1.276 & 0.716 & 1.055 & 0.689 & & 1.351 & 0.817 & 0.923 & 0.688 \\
 & \textbf{$MOMENT_{TikTok}$} & 1.390 & 0.825 & 1.038 & 0.712 & & 1.281 & 0.707 & 0.902 & 0.656 \\
 & \textbf{$MOMENT_{Twitter,TikTok}$} & \textbf{1.110} & \textbf{0.621} & \textbf{0.892} & \textbf{0.651} & & \textbf{1.089} & \textbf{0.720} & \textbf{0.803} & \textbf{0.633} \\
\midrule

\multirow{3}{*}{ Long-term (256)} & \textbf{$MOMENT_{Twitter}$} & 1.283 & 0.714 & 1.070 & 0.693 & & 1.339 & 0.823 & 0.923 & 0.700 \\
 & \textbf{$MOMENT_{TikTok}$} & 1.400 & 0.820 & 1.035 & 0.726 & & 1.250 & 0.728 & 0.916 & 0.674 \\
 & \textbf{$MOMENT_{Twitter,TikTok}$} & \textbf{1.124} & \textbf{0.629} & \textbf{0.907} & \textbf{0.668} & & \textbf{1.091} & \textbf{0.737} & \textbf{0.813} & \textbf{0.654} \\
\bottomrule
\end{tabular}
\label{tab:forecasting}
}
\end{table*}

\subsection{Case Study}
\begin{figure}[htbp]
    \centering
    \begin{subfigure}[b]{0.48\linewidth}
        \centering
        \includegraphics[width=\linewidth]{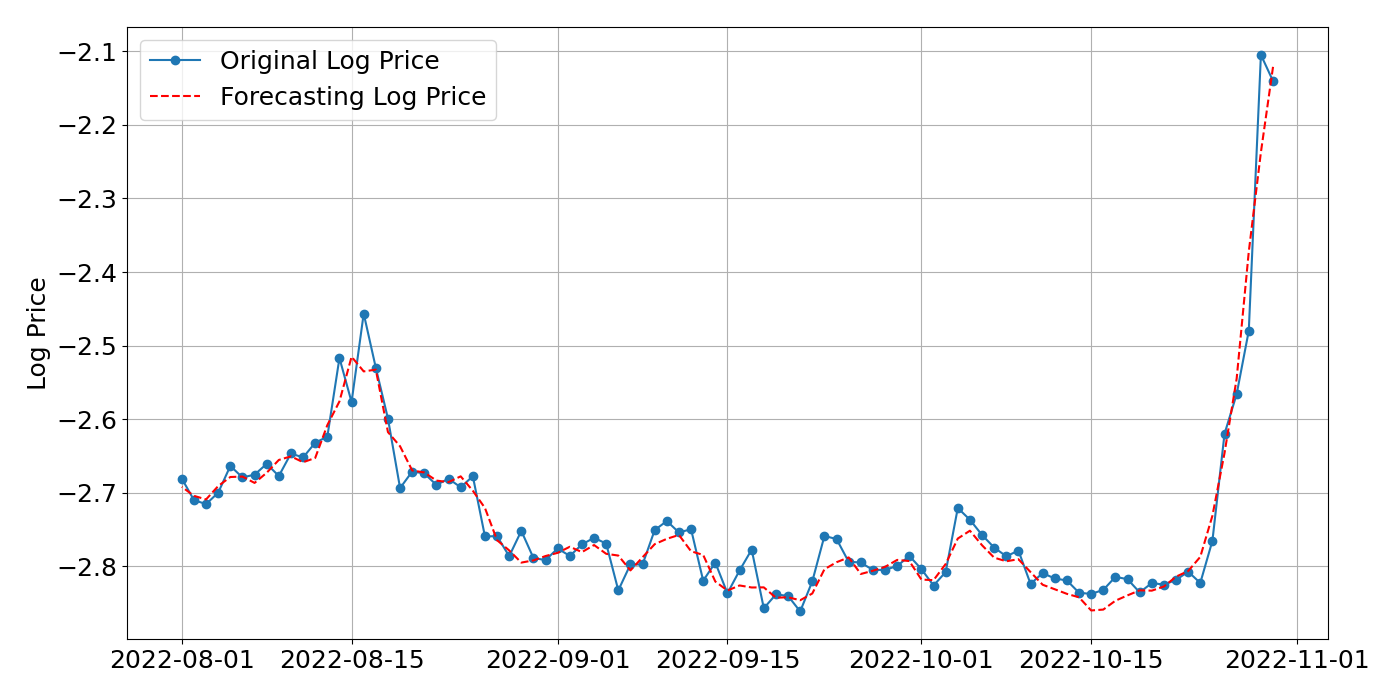}
        \caption{Prediction on DOG price with TikTok}
        \label{fig:fe_dog_tiktok}
    \end{subfigure}
    \hfill
    \begin{subfigure}[b]{0.48\linewidth}
        \centering
        \includegraphics[width=\linewidth]{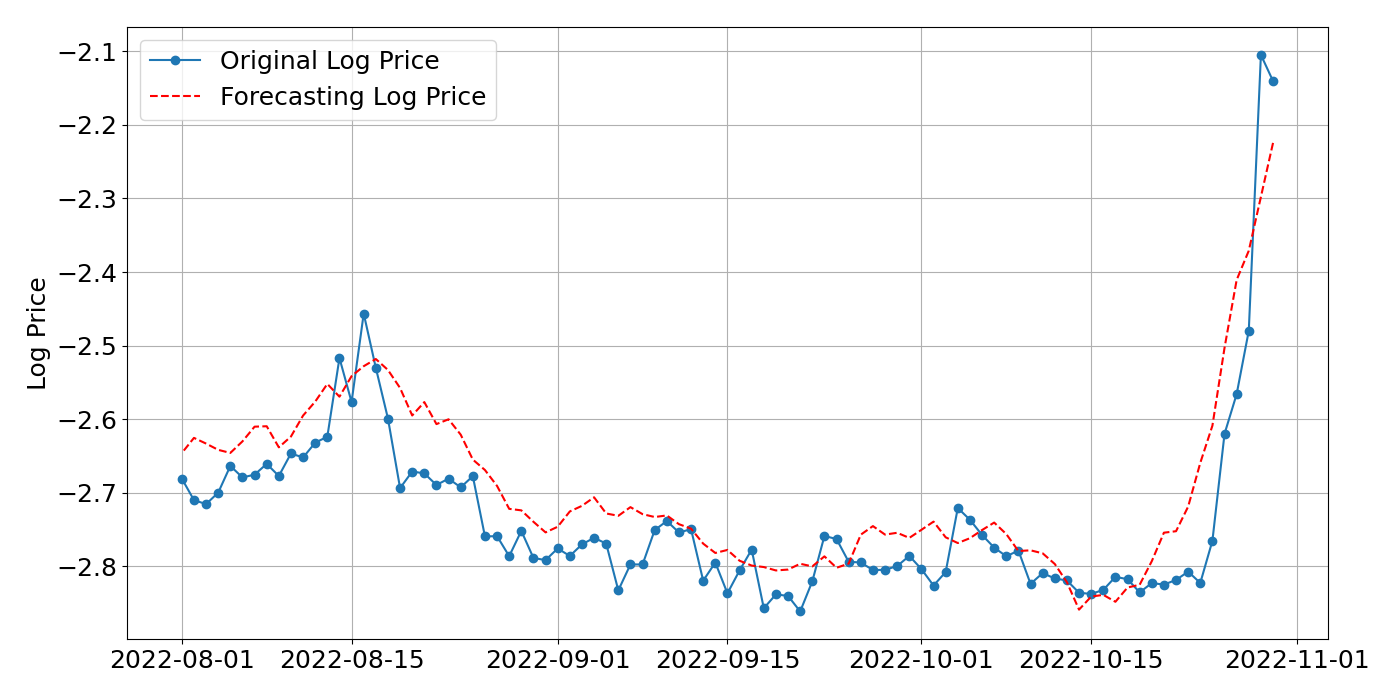}
        \caption{Prediction on DOG price with Twitter}
        \label{fig:fe_dog_twitter}
    \end{subfigure}

    \caption{Predictions of DOG price with sentiment signals from TikTok and Twitter.}
    \label{fig:fe_dog_sentiment}
\end{figure}

The figures illustrate predictions for Dogecoin prices using TikTok (Figure \ref{fig:fe_dog_tiktok}) and Twitter (Figure \ref{fig:fe_dog_twitter}) sentiment score as input signals. Both sets of predictions indicate that sentiments from TikTok and Twitter can follow Dogecoin’s price trends. However, significant differences emerge during periods of rapid price increases, such as in mid-October 2022. These variations highlight the importance of platform-specific sentiments in reflecting market behavior. This suggests that TikTok, with its younger and trend-sensitive user base, may be more effective in capturing speculative price surges. For traders, the results imply that incorporating TikTok sentiment could be useful for monitoring short-term price fluctuations. For instance, speculative investors might rely more heavily on TikTok sentiment to identify quick price movements, while institutional or long-term investors may benefit from using both TikTok and Twitter sentiments to gain a broader understanding of market dynamics.

\section{Conclusion}
\label{sec5}
This paper builds a framework that integrates Twitter and TikTok sentiment to analyze investor mood and cryptocurrency behavior. By combining traditional and multimodal sentiment methods, we captured TikTok’s rapid sentiment shifts and Twitter’s longer-term discussions. Results highlight their complementary roles: TikTok sentiment strongly predicts short-term speculative assets like Dogecoin, while Twitter aligns better with Bitcoin’s medium- to long-term trends. Additionally, merging both platforms consistently improved forecasting accuracy, clarifying social media’s influence on cryptocurrency markets. Future research could integrate additional social media platforms, news sources, and blockchain metrics, alongside macroeconomic factors and regulatory events, to provide a more comprehensive understanding of cryptocurrency market dynamics and reduce platform-specific biases.








\end{document}